\documentclass[11pt,a4paper]{article}
\usepackage[hyperref]{acl2020}
\usepackage{times}
\usepackage{latexsym}

\usepackage{graphicx}  
\usepackage{mathtools}  
\usepackage{amssymb}  
\usepackage{amsthm}  
\usepackage{multirow}  
\usepackage{subfigure}  
\usepackage{booktabs}  
\usepackage{algorithm}  
\usepackage{algorithmic}  
\usepackage{bm}  
\usepackage{xcolor}  
\newcommand{\red}[1]{\textcolor{red}{#1}}  
\newcommand{\itm}[1]{\textrm{\textit{#1}}}  
\newcommand{\bz}{\bm{z}}  
\usepackage{eqparbox}  

\usepackage{microtype}

\aclfinalcopy 
\def\aclpaperid{432} 

\title{On the Encoder-Decoder Incompatibility in\\ Variational Text Modeling and Beyond}

\author{Chen Wu\textsuperscript{1}, Prince Zizhuang Wang\textsuperscript{2}, William Yang Wang\textsuperscript{2} \\
\textsuperscript{1}Department of Foreign Languages and Literatures, Tsinghua University \\
\textsuperscript{2}Department of Computer Science, University of California, Santa Barbara \\
\texttt{wu-c16@mails.tsinghua.edu.cn,} \\
\texttt{zizhuang\_wang@ucsb.edu, william@cs.ucsb.edu} \\
}

\date{}

\begin{document}
\maketitle
\begin{abstract}
Variational autoencoders (VAEs) combine latent variables with amortized variational inference, whose optimization usually converges into a trivial local optimum termed \textit{posterior collapse}, especially in text modeling. By tracking the optimization dynamics, we observe the \textit{encoder-decoder incompatibility} that leads to poor parameterizations of the data manifold. We argue that the trivial local optimum may be avoided by improving the encoder and decoder parameterizations since the posterior network is part of a transition map between them. To this end, we propose Coupled-VAE, which couples a VAE model with a deterministic autoencoder with the same structure and improves the encoder and decoder parameterizations via encoder weight sharing and decoder signal matching. We apply the proposed Coupled-VAE approach to various VAE models with different regularization, posterior family, decoder structure, and optimization strategy. Experiments on benchmark datasets (i.e., PTB, Yelp, and Yahoo) show consistently improved results in terms of probability estimation and richness of the latent space. We also generalize our method to conditional language modeling and propose Coupled-\textbf{C}VAE, which largely improves the diversity of dialogue generation on the Switchboard dataset.\footnote{Our code is publicly available at \url{https://github.com/ChenWu98/Coupled-VAE}.}
\end{abstract}

\section{Introduction}
\label{sec:introduction}
The variational autoencoder (VAE) \cite{KingmaW13} is a generative model that combines neural latent variables and amortized variational inference, which is efficient in estimating and sampling from the data distribution. It infers a posterior distribution for each instance with a shared inference network and optimizes the evidence lower bound (ELBO) instead of the intractable marginal log-likelihood. Given its potential to learn representations from massive text data, there has been much interest in using VAE for text modeling \cite{ZhaoZE17,XuD18,HeSNB19}. 

Prior work has observed that the optimization of VAE suffers from the \textit{posterior collapse} problem, i.e., the posterior becomes nearly identical to the prior and the decoder degenerate into a standard language model \cite{BowmanVVDJB16,ZhaoZE17}. A widely mentioned explanation is that a strong decoder makes the collapsed posterior a good local optimum of ELBO, and existing solutions include weakened decoders \cite{YangHSB17,SemeniutaSB17}, modified regularization terms \cite{HigginsMPBGBML17,WangW19}, alternative posterior families \cite{RezendeM15,DavidsonFCKT18}, richer prior distributions \cite{TomczakW18}, improved optimization strategies \cite{HeSNB19}, and narrowed amortization gaps \cite{KimWMSR18}.

In this paper, we provide a novel perspective for the posterior collapse problem. By comparing the optimization dynamics of VAE with deterministic autoencoders (DAE), we observe the incompatibility between a poorly optimized encoder and a decoder with too strong expressiveness. From the perspective of differential geometry, we show that this issue indicates poor \textit{chart maps} from the data manifold to the \textit{parameterizations}, which makes it difficult to learn a \textit{transition map} between them. Since the posterior network is a part of the transition map, we argue that the posterior collapse would be mitigated with better parameterizations. 

To this end, we propose the Coupled-VAE approach, which couples the VAE model with a deterministic network with the same structure. For better encoder parameterization, we share the encoder weights between the coupled networks. For better decoder parameterization, we propose a signal matching loss that pushes the stochastic decoding signals to the deterministic ones. Notably, our approach is model-agnostic since it does not make any assumption on the regularization term, the posterior family, the decoder architecture, or the optimization strategy. Experiments on PTB, Yelp, and Yahoo show that our method consistently improves the performance of various VAE models in terms of probability estimation and the richness of the latent space. The generalization to conditional modeling, i.e., Coupled-\textbf{C}VAE, largely improves the diversity of dialogue generation on the Switchboard dataset.
Our contributions are as follows: 
\begin{itemize}
    \item We observe the encoder-decoder incompatibility in VAE and connect it to the posterior collapse problem.
    \item We propose the Coupled-VAE, which helps the encoder and the decoder to learn better parameterizations of the data manifold with a coupled deterministic network, via encoder weight sharing and decoder signal matching. 
    \item Experiments on PTB, Yelp, and Yahoo show that our approach improves the performance of various VAE models in terms of probability estimation and richness of the latent space. We also generalize Coupled-VAE to conditional modeling and propose Coupled-\textbf{C}VAE, which largely improves the diversity of dialogue generation on the Switchboard dataset.
\end{itemize}

\section{Background}
\label{sec:background}
\subsection{Variational Inference for Text Modeling}
\begin{figure}[t]
\centering
    \subfigure[Variational autoencoder]{
        \includegraphics[width=0.5\linewidth]{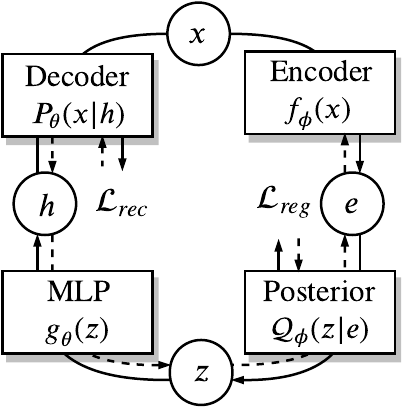}
        \label{subfig:vae}
    }
    \subfigure[Deterministic autoencoder]{
        \includegraphics[width=0.5\linewidth]{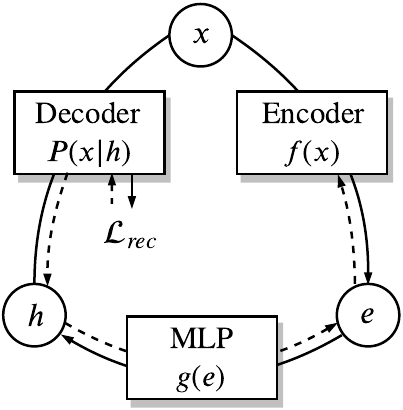}
        \label{subfig:dae}
    }
\caption{VAE and DAE for text modeling.}
\end{figure}

The generative process of VAE is first to sample a latent code $\bz$ from the prior distribution $\mathcal{P}(\bz)$ and then to sample the data $x$ from $P(x|\bz;\theta)$ \cite{KingmaB14}.
Since the exact marginalization of the log-likelihood is intractable, a variational family of posterior distributions $\mathcal{Q}(\bz|x; \phi)$ is adopted to derive the evidence lower bound (ELBO), i.e.,
\begin{equation}
\label{eq:vae-elbo}
\begin{split}
    \log P(x;\theta) &\geq \mathbb{E}_{\bz \sim \mathcal{Q}(\bz|x;\phi)}[\log P(x|\bz;\theta)] \\
    &\ \ \ \ \ - \mathrm{KL}[\mathcal{Q}(\bz|x; \phi) \parallel \mathcal{P}(\bz)]
\end{split}
\end{equation}
For training, as shown in Figure~\ref{subfig:vae}, the encoded text $\bm{e}$ is transformed into its posterior via a posterior network. A latent code is sampled and mapped to the decoding signal $\bm{h}$. Finally, the decoder infers the input with the decoding signal. The objective can be viewed as a reconstruction loss $\mathcal{L}_{\itm{rec}}$ plus a regularization loss $\mathcal{L}_{\itm{reg}}$ (whose form varies), i.e.,
\begin{equation}
\label{eq:vae-general-loss}
    \mathcal{L} = \mathcal{L}_{\itm{rec}} + \mathcal{L}_{\itm{reg}}
\end{equation}
However, the optimization of the VAE objective is challenging. We usually observe a very small $\mathcal{L}_{\itm{reg}}$ and a $\mathcal{L}_{\itm{rec}}$ similar to a standard language model, i.e., the well-known \textit{posterior collapse} problem. 

\subsection{Deterministic Autoencoders}
An older family of autoencoders is the deterministic autoencoder (DAE) \cite{rumelhart1986learning,Ballard87}. Figure~\ref{subfig:dae} shows an overview of DAE for text modeling, which is composed of a text encoder, an optional MLP, and a text decoder. The reconstruction loss of DAE is usually much lower than that of VAE after convergence.

\section{Encoder-Decoder Incompatibility in VAE for Text Modeling}
\label{sec:analysis}
To understand the posterior collapse problem, we take a deeper look into the training dynamics of VAE. We investigate the following questions. How much backpropagated gradient does the encoder receive from reconstruction? How much does it receive from regularization? How much information does the decoder receive from the encoded text?

\subsection{Tracking Training Dynamics}
\label{subsec:tracking}
\begin{figure*}[t]
\centering
    \subfigure[Gradient norm of the reconstruction loss w.r.t. the encoded text]{
        \includegraphics[width=0.3\linewidth]{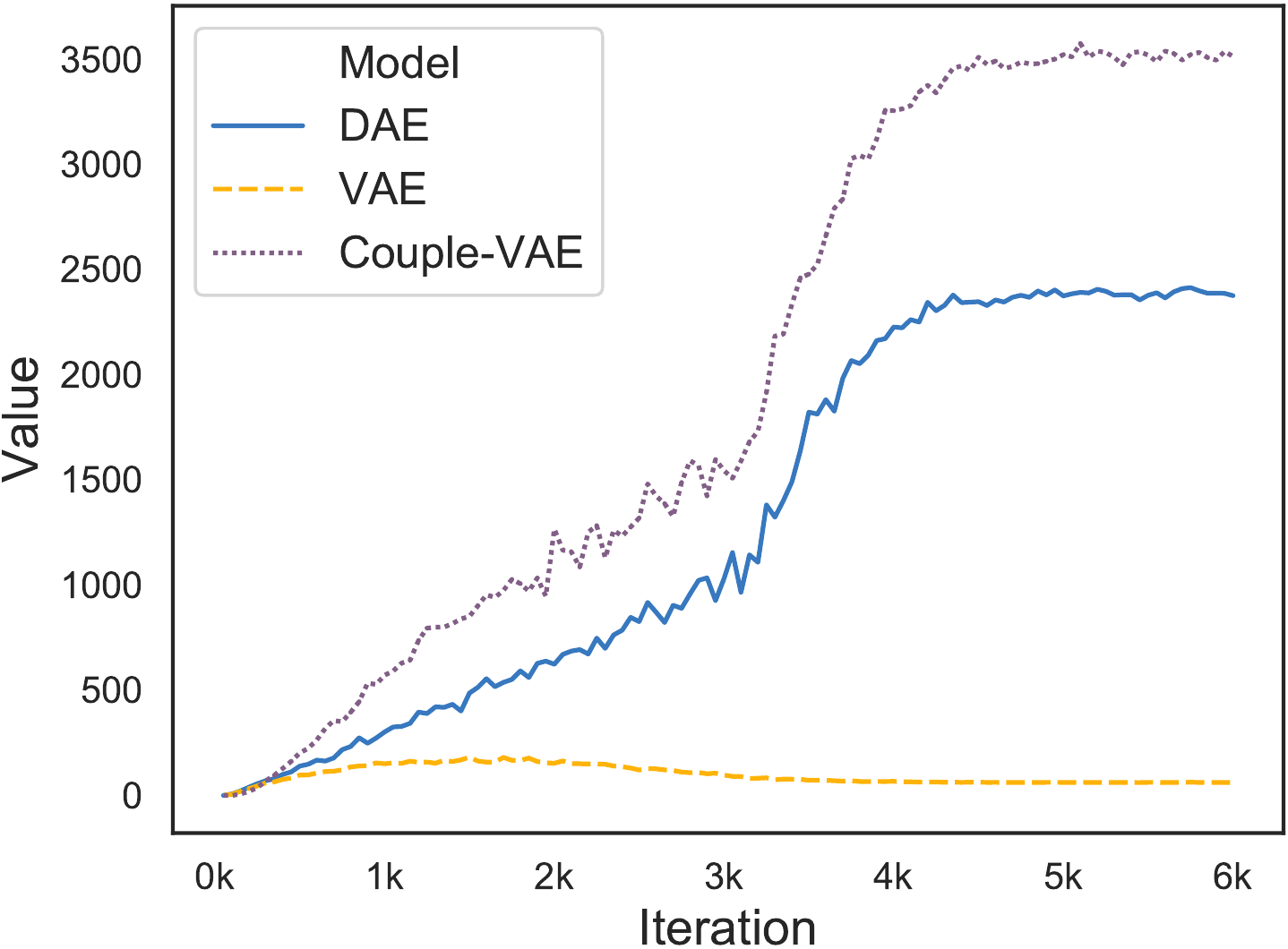}
        \label{subfig:curve-derecde}
    }
\hspace{0.02\linewidth}
    \subfigure[Gradient norm of the regularization loss w.r.t. the encoded text]{
        \includegraphics[width=0.3\linewidth]{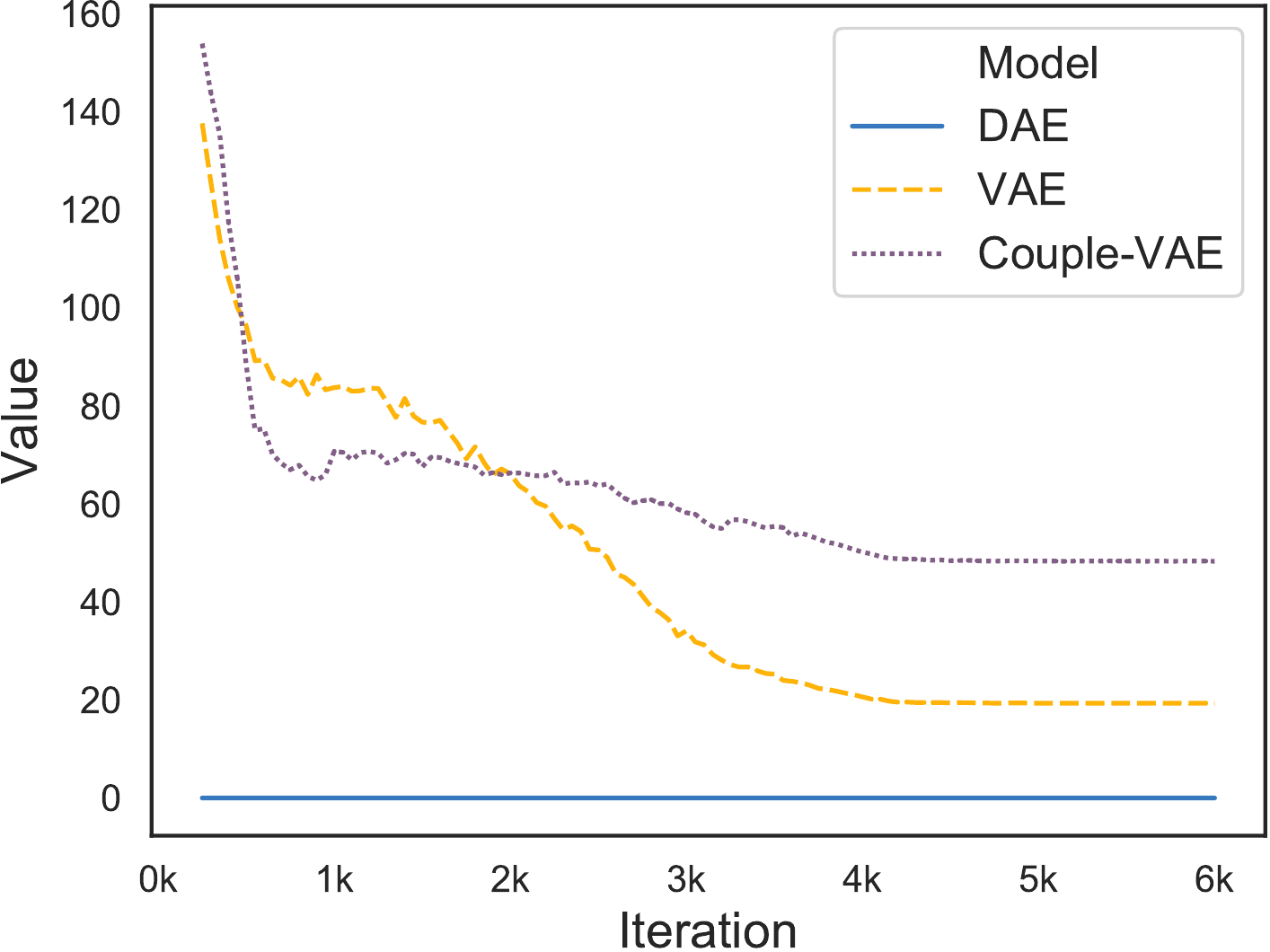}
        \label{subfig:cuurve-dregde}
    }
\hspace{0.02\linewidth}
    \subfigure[Gradient norm of the decoding signal w.r.t. the encoded text]{
        \includegraphics[width=0.3\linewidth]{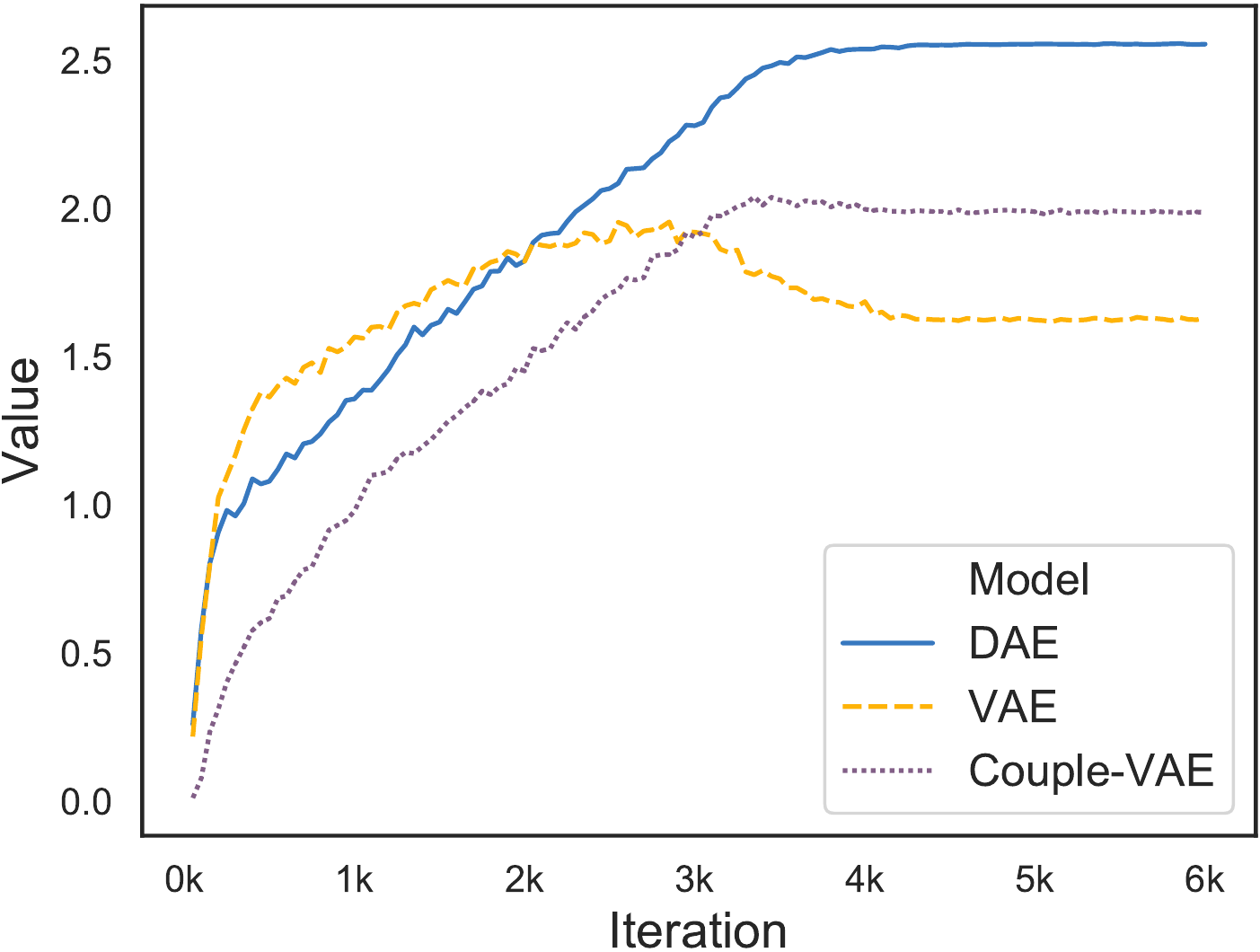}
        \label{subfig:cuurve-dhde}
    }
\caption{\label{fig:gradient-norms-curve} Training dynamics of DAE, VAE, and the proposed Coupled-VAE on the Yelp test set. Please find the analysis in Section~\ref{sec:analysis} and Section~\ref{subsec:analysis-gradients}. Best viewed in color (yet the models are distinguished by line markers).}
\end{figure*}
To answer the first question, we study the gradient norm of the reconstruction loss w.r.t. the encoded text, i.e., $\left\| \partial \mathcal{L}_{\itm{rec}} / \partial \bm{e} \right\|_{2}$, which shows the magnitude of gradients received by the encoder parameters. From Figure~\ref{subfig:curve-derecde}, we observe that it constantly increases in DAE, while in VAE it increases marginally in the early stage and then decreases continuously. It shows that the reconstruction loss actively optimizes the DAE encoder, while the VAE encoder lacks backpropagated gradients after the early stage of training.

We seek the answer to the second question by studying the gradient norm of the regularization loss w.r.t. the encoded text, i.e., $\left\| \partial \mathcal{L}_{\itm{reg}} / \partial \bm{e} \right\|_{2}$. In a \textit{totally} collapsed posterior, i.e., $\mathcal{Q}(\bz|x; \phi) = \mathcal{P}(\bz)$ for each $x$, $\left\| \partial \mathcal{L}_{\itm{reg}} / \partial \bm{e} \right\|_{2}$ would be zero. Thus, $\left\| \partial \mathcal{L}_{\itm{reg}} / \partial \bm{e} \right\|_{2}$ can show how far the posterior of each instance is from the aggregate posterior or the prior. Figure~\ref{subfig:cuurve-dregde} shows a constant decrease of the gradient norm in VAE from the 2.5K step until convergence, which shows that the posterior collapse is aggravated as the KL weight increases.  

For the third question, we compute the normalized gradient norm of the decoding signal w.r.t. the encoded text, i.e., $\left\| \partial \bm{h} / \partial \bm{e} \right\|_{F} / \left\| \bm{h} \right\|_{2}$. As this term shows how relatively the decoding signal changes with the perturbation of the encoded text, it reflects the amount of information passed from the encoder to the decoder. Figure~\ref{subfig:cuurve-dhde} shows that for DAE, it constantly increases. For VAE, it at first increases even faster than DAE, slows down, and finally decreases until convergence, indicating that the VAE decoder, to some extent, ignores the encoder in the late stage of training.

\begin{figure*}[t]
    \centering
    \subfigure{
        \includegraphics[width=0.27\linewidth]{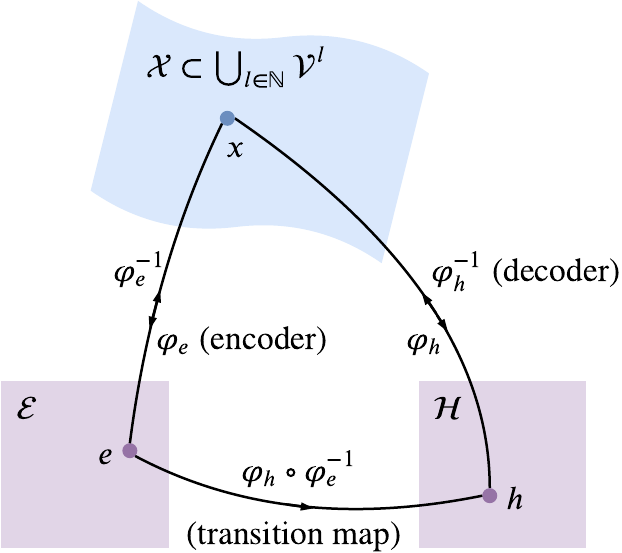}
    }
    \hspace{0.03\linewidth}
    \subfigure{
        \includegraphics[width=0.55\linewidth]{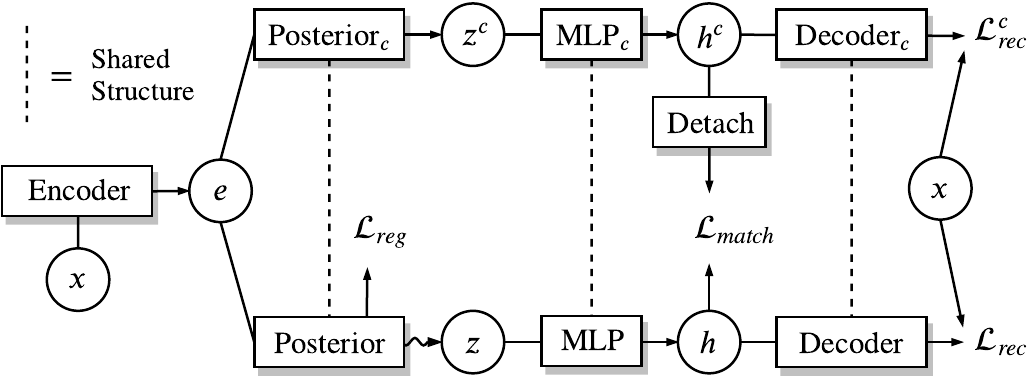}
    }
    \caption{\label{fig:Coupled-VAE} Left: DAE and VAE interpreted as manifold parameterizations and a transition map. Right: A graphical overview of the proposed Coupled-VAE. The upper path is deterministic, and the lower path is stochastic.}
\end{figure*}

\subsection{Encoder-Decoder Incompatibility}
\label{subsec:incompatibility}
Based on the training dynamics in Section~\ref{subsec:tracking} and the observations in previous work \cite{BowmanVVDJB16,ZhaoZE17}, text VAE has three features, listed as follows. First, the encoder is poorly optimized, as shown by the low $\left\| \partial \mathcal{L}_{\itm{rec}} / \partial \bm{e} \right\|_{2}$. Second, the decoder degenerates into a powerful language model. Third, $\bm{h}$ contains less information from $\bm{e}$ in VAE than in DAE, which is indicated by the lower $\left\| \partial \bm{h} / \partial \bm{e} \right\|_{F} / \left\| \bm{h} \right\|_{2}$. We call these features as \textit{encoder-decoder incompatibility}.

To bridge the incompatibility and posterior collapse, we start with the \textit{manifold hypothesis} which states that real-world data concentrates near a manifold with a lower dimensionality than the ambient space \cite{NarayananM10,BengioCV13}. In our case, we denote the manifold of text data as $\mathcal{X} \subset \bigcup_{l \in \mathbb{N}} \mathcal{V}^{l}$ where $\mathcal{V}$ is the vocabulary. In the language of differential geometry, the encoded text $\bm{e} \in \mathcal{E} \subset \mathbb{R}^{d}$ and the decoding signal $\bm{h} \in \mathcal{H} \subset \mathbb{R}^{d}$ can be viewed as the \textit{parameterizations} (or \textit{coordinates}) of $x \in \mathcal{X}$ under two different \textit{charts} (or \textit{coordinate systems}). Formally, we denote the chart maps as $\varphi_{e}: \mathcal{X} \rightarrow \mathcal{E}$ and $\varphi_{h}: \mathcal{X} \rightarrow \mathcal{H}$, which satisfy $\bm{e} = \varphi_{e}(x)$ and $\bm{h} = \varphi_{h}(x)$ for any $x \in \mathcal{X}$. Given the two charts, the map from $\mathcal{E}$ to $\mathcal{H}$ is called the \textit{transition map} $\varphi_{h} \circ \varphi_{e}^{-1}: \mathcal{E} \rightarrow \mathcal{H}$ between the two charts. 

In DAE, the two chart maps and the transition map between them are learned simultaneously via the single reconstruction loss, which we rewrite as
\begin{equation}
\label{eq:diff-geom-rec-loss}
    \mathcal{L}_{\itm{rec}} = \mathbb{E}_{x \in \mathcal{X}}[\mathcal{L}(x, \varphi_{h}^{-1}(\varphi_{h} \circ \varphi_{e}^{-1}(\varphi_{e}(x))))]
\end{equation}
where $\varphi_{e}$, $\varphi_{h} \circ \varphi_{e}^{-1}$, and $\varphi_{h}^{-1}$ are modeled as the encoder, the MLP, and the decoder (strictly speaking, in text modeling, the range of $\varphi_{h}^{-1}$ is not $\mathcal{X}$ but distributions on $\mathcal{X}$), as  illustrated in Figure~\ref{fig:Coupled-VAE}.

In VAE, as discussed before, both $\varphi_{e}$ and $\varphi_{h}$ inadequately parameterize the data manifold. We argue that the inadequate parameterizations make it harder to find a smooth transition map in VAE than in DAE, as shown by the lower $\left\| \partial \bm{h} / \partial \bm{e} \right\|_{F} / \left\| \bm{h} \right\|_{2}$. Since the posterior network is a part of the transition map, it consequently seeks to map each instance to the prior (discussed in Section~\ref{subsec:tracking}) rather than learning the transition map.

\section{Coupling Variational and Deterministic Networks}
\label{sec:methodology}
Based on the above analysis, we argue that posterior collapse could be alleviated by learning chart maps (i.e., $\varphi_{e}$ and $\varphi_{h}$) that better parameterize the data manifold. Inspired by the chart maps in DAE, we propose to couple the VAE model with a deterministic network, outlined in Figure~\ref{fig:Coupled-VAE}. Modules with a subscript $c$ are deterministic networks that share the \textit{structure} with those in the stochastic network.
Sampling is disabled in the deterministic network, e.g., in the case of Gaussian posterior, we use the predicted mean vector for later computation. Please find details for other posterior families in Appendix~\ref{app:deterministic-details}. Similar to DAE, the coupled deterministic network is optimized solely by the coupled reconstruction loss $\mathcal{L}_{\itm{rec}}^{c}$, which is the same autoregressive cross-entropy loss as $\mathcal{L}_{\itm{rec}}$. 

To learn a well-optimized $\varphi_{e}$, we share the encoder between the stochastic and the deterministic networks, which leverages the rich gradients backpropagated from $\mathcal{L}_{\itm{rec}}^{c}$. To learn better $\varphi_{h}$, we propose to guide $\varphi_{h}$ with a well-learned chart map, i.e., the one characterized by $\mathrm{Decoder}^c$. Thus, we introduce a signal matching loss $\mathcal{L}_{\itm{match}}$ that pushes the $\bm{h}$ to $\bm{h}^{c}$. The objective of our approach is
\begin{equation}
\label{eq:Coupled-VAE}
    \mathcal{L} = \mathcal{L}_{\itm{rec}} + \mathcal{L}_{\itm{reg}} + \lambda_{r} \mathcal{L}_{\itm{rec}}^{c} + \lambda_{m} \mathcal{L}_{\itm{match}}
\end{equation}
where $\lambda_{r}$ and $\lambda_{m}$ are hyperparameters\footnote{To avoid heavy hyperparameter tuning, we set $\lambda_{r}=1.0$ unless otherwise specified.}, $\mathcal{L}_{\itm{rec}}^{c}$ is the coupled reconstruction loss, and the signal matching loss $\mathcal{L}_{\itm{match}}$ is essentially a distance function between $\bm{h}$ and $\bm{h}^{c}$. We evaluate both the Euclidean distance and the Rational Quadratic kernel\footnote{To avoid heavy hyperparameter tuning, we use the Rational Quadratic kernel unless otherwise specified.}, i.e., 
\begin{equation}
\label{eq:l-couple}
    \mathcal{L}_{\itm{match}} = \left\{
    \begin{array}{rcl}
        \left\| \bm{h} - \mathrm{Detach(}\bm{h}^{c}\mathrm{)} \right\|^{2} \hfill \ \ \ \textrm{Eucl} \\
        \sum_{s}\frac{- s \cdot C}{s \cdot C + \left\| \bm{h} - \mathrm{Detach(}\bm{h}^{c}\mathrm{)} \right\|^{2}} \hfill \ \ \ \textrm{RQ}
    \end{array} \right.
\end{equation}
where $s \in \{0.1, 0.2, 0.5, 1, 2, 5, 10\}$, $C$ is a hyper-parameter, and $\mathrm{Detach}$ prevents gradients to be propagated into $\bm{h}^{c}$ since we would like $\bm{h}^{c}$ to guide $\bm{h}$ but not the opposite.

One would question the necessity of sharing the structure of the posterior network by resorting to universal approximation \cite{HornikSW89}. Specifically, a common question is: why not using an MLP as $\textrm{Posterior}_{c}$? We argue that each structure has a favored distribution of $\mathcal{H}$ in $\mathbb{R}^{d}$, so structure sharing facilitates the optimization when we are learning by gradient descent. For example, the latent space learned by planar flows \cite{RezendeM15} has compression and expansion, and vMF-VAE \cite{XuD18}, which is supported on a sphere, may significantly influence the distribution of $\mathcal{H}$ in its ambient space $\mathbb{R}^{d}$.

\begin{table*}[t]
\centering
\small
\begin{tabular}{@{}llrlrlr@{}}
\toprule
\multirow{2}*{} & \multicolumn{2}{c}{PTB} & \multicolumn{2}{c}{Yelp} & \multicolumn{2}{c}{Yahoo}\\
\cmidrule(lr){2-3} \cmidrule(lr){4-5} \cmidrule(l){6-7}
~                                       & NLL (KL)          & PPL       & NLL (KL)          & PPL       & NLL (KL)          & PPL \\
\midrule
GRU-LM*                                 & 105.8 (-)         & 125.3     & 196.3 (-)         & 57.3      & 347.9 (-)         & 78.0 \\
\midrule
VAE                                     & 103.6 (8.6)       & 112.9     & 193.7 (7.2)       & 54.3      & 344.5 (12.4)      & 74.7 \\
Coupled-VAE                              & \bf 103.1 (9.5)   & \bf 110.5 & \bf 191.2 (8.0)   & \bf 51.6  & \bf 342.4 (12.8)  & \bf 72.8 \\
\midrule
$\beta(0.8)$-VAE                        & 103.8 (11.0)      & 113.9     & 193.8 (10.2)      & 54.5      & 344.9 (16.1)      & 75.1 \\
Coupled-$\beta(0.8)$-VAE                 & \bf 103.3 (12.1)  & \bf 111.5 & \bf 191.5 (12.2)  & \bf 51.9  & \bf 342.8 (17.0)  & \bf 73.2 \\
\midrule
$\beta(1.2)$-VAE                        & 103.7 (7.8)       & 113.3     & 193.7 (6.0)       & 54.3      & 345.3 (10.5)      & 75.5 \\
Coupled-$\beta(1.2)$-VAE                 & \bf 102.9 (8.6)   & \bf 109.6 & \bf 191.2 (6.9)   & \bf 51.6  & \bf 342.3 (11.3)  & \bf 72.7  \\
\midrule
vMF-VAE                                 & 103.6 (2.0)       & 113.2     & 195.4 (0.0)       & 56.3      & 344.5 (2.5)       & 74.7 \\
Coupled-vMF-VAE                          & \bf 103.0 (3.0)   & \bf 110.1 & \bf 191.2 (2.8)   & \bf 51.6  & \bf 342.2 (4.0)   & \bf 72.5 \\
\midrule
CNN-VAE                                 & 118.5 (29.6)      & 222.6     & 194.2 (12.8)      & 54.8      & 344.3 (19.7)      & 74.5 \\
Coupled-CNN-VAE                          & \bf 118.2 (30.2)  & \bf 219.7 & \bf 193.9 (13.7)  & \bf 54.6  & \bf 343.3 (22.4)   & \bf 73.6 \\
\midrule
WAE                                     & 103.7 (11.0)      & 113.3     & 193.7 (10.7)      & 54.3      & 344.7 (16.6)      & 74.9 \\
Coupled-WAE                              & \bf 103.2 (12.5)  & \bf 110.9 & \bf 191.3 (12.5)  & \bf 51.7  & \bf 343.3 (18.2)  & \bf 73.6 \\
\midrule
VAE-NF                                  & 103.3 (5.5)       & 111.3     & 193.9 (5.3)       & 54.5      & 344.3 (8.1)       & 74.5 \\
Coupled-VAE-NF                           & \bf 102.6 (5.7)   & \bf 108.1 & \bf 191.8 (5.6)   & \bf 52.2  & \bf 342.6 (8.8)   & \bf 73.0 \\
\midrule
WAE-NF                                  & 103.4 (6.7)       & 111.9     & 194.1 (7.0)       & 54.7      & 344.3 (10.6)      & 74.5 \\
Coupled-WAE-NF                           & \bf 102.7 (7.4)   & \bf 108.4 & \bf 192.1 (7.4)   & \bf 52.5  & \bf 342.7 (11.0)  & \bf 73.1 \\
\midrule
CycAnn-VAE                              & 104.2 (1.6)       & 116.3     & 192.5 (1.2)       & 53.0      & 345.4 (3.9)       & 75.5 \\
Coupled-CycAnn-VAE                       & \bf 103.7 (2.4)   & \bf 113.3 & \bf 190.8 (2.0)   & \bf 51.1  & \bf 342.4 (4.4)   & \bf 72.7 \\
\midrule
PreFB-VAE                               & 103.4 (14.6)      & 111.9     & 190.4 (14.1)      & 50.7      & 341.4 (17.6)      & 71.8 \\
Coupled-PreFB-VAE                        & \bf 103.3 (15.6)  & \bf 111.4 & \bf 189.9 (14.4)  & \bf 50.3  & \bf 341.3 (17.9)  & \bf 71.7 \\
\midrule
SA-VAE$^{\dag}$                         & 100.7 (7.7)       & 98.7      & 183.5 (3.8)       & 44.0      & 327.5 (7.2)$^\ddag$ & 60.4$^\ddag$ \\
\midrule
Lagging-VAE$^{\dag}$                    & 98.8 (6.0)        & 90.7      & 182.5 (1.2)       & 43.1      & 326.7 (6.0)       & 59.7 \\
Coupled-Lagging-VAE$^{\dag}$             & \bf 98.7 (11.0)   & \bf 90.4  & \bf 182.3 (3.8)   & \bf 42.9  & \bf 326.2 (7.4)   & \bf 59.3 \\
\bottomrule
\end{tabular}
\caption{\label{tab:lm-results} Language modeling results. NLL is estimated with importance sampling. PPL is based on the estimated NLL. KL and MI are approximated by their Monte Carlo estimates. \textit{Coupled-} stands for ``with the coupled deterministic network''. The better results in each block are shown in \textbf{bold}. *The exact NLL is reported. $^\dag$Modifying open-source implementation which does not follow our setup and evaluation. $^\ddag$Previously reported.}
\end{table*}

\section{Experiments}
\label{sec:experiments}
\subsection{Datasets}
We conduct the experiments on three commonly used datasets for text modeling, i.e., the Penn Treebank (PTB) \cite{MarcusSM94}, Yelp \cite{XuCQH16}, and Yahoo. The training/validation/test splits are 42K/3370/3761 for PTB, 63K/7773/8671 for Yelp, and 100K/10K/10K for Yahoo. The vocabulary size for PTB/Yelp/Yahoo is 10K/15K/20K. We discard the sentiment labels in Yelp. 

\subsection{Baselines}
We evaluate the proposed Coupled-VAE approach by applying it to various VAE models, which include VAE \cite{KingmaW13}, $\beta$-VAE \cite{HigginsMPBGBML17}, vMF-VAE \cite{XuD18,DavidsonFCKT18} with learnable $\kappa$, CNN-VAE \cite{YangHSB17}, WAE \cite{TolstikhinBGS18}, VAE with normalizing flows (VAE-NF) \cite{RezendeM15}, WAE with normalizing flows (WAE-NF), VAE with cyclic annealing schedule (CycAnn-VAE) \cite{FuLLGCC19}, VAE with encoder pretraining and the free bits objective (PreFB-VAE) \cite{LiHNBY19}, and Lagging-VAE \cite{HeSNB19}. We also show the result of GRU-LM \cite{ChoMGBBSB14} and SA-VAE \cite{KimWMSR18}. We do not apply our method to SA-VAE since it does not follow amortized variational inference. Please find more details in Appendix~\ref{app:setup} and previous footnotes. 

\begin{table*}[t]
\centering
\small
\begin{tabular}{@{}lrrrrrr@{}}
\toprule
\multirow{2}*{} & \multicolumn{2}{c}{PTB} & \multicolumn{2}{c}{Yelp} & \multicolumn{2}{c}{Yahoo}\\
\cmidrule(lr){2-3} \cmidrule(lr){4-5} \cmidrule(l){6-7}
~                                       & MI \ \    & BLEU-1/2          & MI \ \    & BLEU-1/2          & MI \ \    & BLEU-1/2 \\
\midrule
VAE                                     & 10.48     & 23.2 / 4.4        & 8.28      & 28.7 / 5.3        & 15.43     & 21.2 / 3.6 \\
Coupled-VAE                              & \bf 11.99 & \bf 23.4 / 4.5    & \bf 9.65  & \bf 30.4 / 5.8    & \bf 16.44 & \bf 23.1 / 4.1 \\
\midrule
$\beta(0.8)$-VAE                        & 15.43     & \bf 24.5 / 4.9    & 13.52     & 30.6 / 6.0        & 24.16     & 24.0 / 4.3 \\
Coupled-$\beta(0.8)$-VAE                 & \bf 18.13 & 24.3 / 4.8        & \bf 17.69 & \bf 32.6 / 6.6    & \bf 28.03 & \bf 26.4 / 4.9 \\
\midrule
$\beta(1.2)$-VAE                        & 9.16      &22.8 / \textbf{4.3}& 6.60      & 28.0 / 5.0        & 11.83     & 18.2 / 2.9 \\
Coupled-$\beta(1.2)$-VAE                 & \bf 10.28 &\textbf{22.9} / 4.2& \bf 7.90  & \bf 29.8 / 5.6    & \bf 13.51 & \bf 22.4 / 3.8 \\
\midrule
vMF-VAE                                 & 1.74      & 15.2 / 2.0        & 0.03      & 22.4 / 2.8        & 2.06      & 8.5 / 1.1 \\
Coupled-vMF-VAE                          & \bf 2.37  & \bf 16.1 / 2.3    & \bf 2.60  & \bf 25.1 / 4.0    & \bf 3.37  & \bf 10.3 / 1.4 \\
\midrule
CNN-VAE                                 & 78.49     & \bf 32.0 / 7.8    & 17.26     & 32.9 / 7.1        & 30.18     & 24.9 / 5.3 \\
Coupled-CNN-VAE                          & \bf 80.54 & 31.8 / 7.7        & \bf 19.15 & \bf 33.4 / 7.3    & \bf 37.62 & \bf 26.9 / 5.9 \\
\midrule
WAE                                     & 15.09     & \bf 24.8 / 5.1    & 15.08     & 30.7 / 6.1        & 24.73     & 24.2 / 4.5 \\
Coupled-WAE                              & \bf 18.51 & 24.7 / \bf 5.1    & \bf 18.56 & \bf 32.5 / 6.6    & \bf 30.08 & \bf 27.7 / 5.3 \\
\midrule
VAE-NF                                  & 5.63      & 19.2 / \bf 3.3    & 5.64      & 25.6 / 4.5        & 8.02      & 13.7 / 2.1 \\
Coupled-VAE-NF                           & \bf 5.86  & \bf 19.4 / 3.3    & \bf 6.06  & \bf 26.3 / 4.6    & \bf 9.14  & \bf 15.3 / 2.5 \\
\midrule
WAE-NF                                  & 7.18      & 19.7 / 3.5        & 7.95      & 26.0 / 4.6        & 11.43     & 13.8 / 2.2 \\
Coupled-WAE-NF                           & \bf 8.10  & \bf 20.7 / 3.7    & \bf 8.53  & \bf 27.2 / 5.0    & \bf 12.56 & \bf 14.9 / 2.5 \\
\midrule
CycAnn-VAE                              & 1.55      & 16.3 / 2.3        & 1.18      & 22.6 / 3.2        & 3.09      & 8.3 / 1.1 \\
Coupled-CycAnn-VAE                       & \bf 2.27  & \bf 16.7 / 2.6    & \bf 2.01  & \bf 23.1 / 3.4    & \bf 3.89  & \bf 10.9 / 1.5 \\
\midrule
PreFB-VAE                               & 20.6      & 25.5 / 5.7        & 20.3      & 33.1 / 6.8        & 26.2      & 27.2 / 5.2 \\
Coupled-PreFB-VAE                        & \bf 23.2  & \bf 25.8 / 5.8    & \bf 21.0  & \bf 33.3 / 6.8    & \bf 27.0  & \bf 27.2 / 5.3 \\
\midrule
Lagging-VAE$^{\dag}$                    & 2.90      & \multicolumn{1}{c}{-} & 0.96      & \multicolumn{1}{c}{-} & 3.04     & \multicolumn{1}{c}{-} \\
Coupled-Lagging-VAE$^{\dag}$             & \bf 3.29  & \multicolumn{1}{c}{-} & \bf 2.36  & \multicolumn{1}{c}{-} & \bf 3.06 & \multicolumn{1}{c}{-} \\
\bottomrule
\end{tabular}
\caption{\label{tab:info-results} Mutual information (MI) and reconstruction. $^{\dag}$Modifying the open-source implementation.}
\end{table*}

\subsection{Language Modeling Results}
\label{subsec:language-modeling}
We report negative log-likelihood (NLL), KL divergence, and perplexity as the metrics for language modeling. NLL is estimated with importance sampling, KL is approximated by its Monte Carlo estimate, and perplexity is computed based on NLL. Please find the metric details in Appendix~\ref{app:lm-metrics}. 

Table~\ref{tab:lm-results} displays the language modeling results. For all models, our proposed approach achieves smaller negative log-likelihood and lower perplexity, which shows the effectiveness of our method to improve the probability estimation capability of various VAE models. Larger KL divergence is also observed, showing that our approach helps address the posterior collapse problem.

\subsection{Mutual Information and Reconstruction}
Language modeling results only evaluate the probability estimation ability of VAE. We are also interested in how rich the latent space is. We report the mutual information (MI) between the text $x$ and the latent code $\bz$ under $\mathcal{Q}(\bz|x)$, which is approximated with Monte Carlo estimation. Better reconstruction from the encoded text is another way to show the richness of the latent space. For each text $x$, we sample ten latent codes from $\mathcal{Q}(\bz|x)$ and decode them with greedy search. We report the BLEU-1 and BLEU-2 scores between the reconstruction and the input. Please find the metric details in Appendix~\ref{app:info-metrics}. In Table~\ref{tab:info-results}, we observe that our approach improves MI on all datasets, showing that our approach helps learn a richer latent space. BLEU-1 and BLEU-2 are consistently improved on Yelp and Yahoo, but not on PTB. Given that text samples in PTB are significantly shorter than those in Yelp and Yahoo, we conjecture that it is easier for the decoder to reconstruct on PTB by exploiting its autoregressive expressiveness, even without a rich latent space.

\begin{table*}[t]
\centering
\small
\begin{tabular}{@{}llllccrlccr@{}}
\toprule
~  & ~   & ~ & \multicolumn{4}{c}{PTB} & \multicolumn{4}{c}{Yelp}\\
\cmidrule(lr){4-7} \cmidrule(l){8-11}
Dist & $\lambda_{m}$    & $\lambda_{r}$             & NLL (KL)          & PPL       & MI        & BLEU-1/2          & NLL (KL)          & PPL       & MI        & BLEU-1/2 \\
\midrule
\multirow{3}*{RQ} & $0.1$* &  \multirow{3}*{$1.0$}   & \bf 103.1 (9.5)   & \bf 110.5 & 11.99     & 23.4 / 4.5        & 191.2 (8.0)       & 51.6      & 9.65      & 30.4 / 5.8 \\
~ &$1.0$   &                                        & 103.3 (10.7)      & 111.4     & 14.32     & 24.0 / 4.8        & \bf 191.1 (8.1)   & \bf 51.5  & 9.92      & 30.5 / 5.8 \\
~ &$5.0$ &                                          & 103.7 (16.1)      & 113.2     & \bf 32.78 & \bf 26.5 / 5.8    & 191.5 (12.8)      & 51.9      & \bf 19.77 & \bf 32.8 / 6.5 \\
 \midrule
\multirow{4}*{RQ} & \multirow{4}*{$0.1$} & $0.0$  & 104.1 (7.3) & 115.3 &  8.60 & 21.0 / 3.7  & 191.7 (5.8) & 52.1 & 6.40 & 27.7 / 5.0 \\
&  & $0.5$    & 103.4 (9.2)       & 111.8     & 11.58     & 23.1 / 4.3        & 191.3 (7.8)       & 51.7      & 9.32      & 29.8 / 5.7  \\
&  & $1.0$*                                          & \bf 103.1 (9.5)   & \bf 110.5 & \bf 11.99 & \bf 23.4 / 4.5    & \bf 191.2 (8.0)   & \bf 51.6  & \bf 9.65  & \bf 30.4 / 5.8 \\
&  & $5.0$                                          & \bf 103.1 (9.1)   & 110.6     & 11.15     & 22.9 / 4.4        & 192.9 (8.0)       & 53.4      & 9.53      & 30.0 / 5.8 \\
\midrule 
\multirow{3}*{Eucl} & $0.1$  & \multirow{3}*{$1.0$} & \bf 103.3 (10.1)  & \bf 111.5 & 13.25     & 23.4 / 4.7        & \bf 191.2 (9.2)   & \bf 51.6  & 11.69     & 31.1 / 6.0 \\
~ &$1.0$ &                                          & 103.9 (17.4)      & 114.5     & 30.52     & 27.7 / 6.1        & 192.1 (14.3)      & 52.5      & 23.14     & 33.8 / 6.9 \\
~ &$5.0$    &                                       & 108.9 (33.3)      & 144.0     & \bf 98.02 & \bf 32.0 / 8.5    & 194.4 (25.0)      & 55.1      & \bf 61.62 & \bf 36.8 / 8.2 \\
\midrule
& VAE &                                             & 103.6 (8.6)       & 112.9 & 10.48 & 23.2 / 4.4    & 193.7 (7.2)   & 54.3  & 8.28  & 28.7 / 5.3 \\
\bottomrule
\end{tabular}
\caption{\label{tab:matching-weight} Hyperparameter analysis. The best results in each block are shown in \textbf{bold}. *Reported in Table~\ref{tab:lm-results} and \ref{tab:info-results}.}
\end{table*}
\subsection{Hyperparameter Analysis: Distance Function, $\lambda_{r}$, and $\lambda_{m}$}
\label{subsec:hyperparameter-analysis}
We investigate the effect of key hyperparameters. Results are shown in Table~\ref{tab:matching-weight}. Note that the lowest NLL does not guarantee the best other metrics, which shows the necessity to use multiple metrics for a more comprehensive evaluation.

For the distance function, we observe that the Euclidean distance (denoted as Eucl in Table~\ref{tab:matching-weight}) is more sensitive to $\lambda_{m}$ than the Rational Quadratic kernel (denoted as RQ in Table~\ref{tab:matching-weight}). 

The first and the third block in Table~\ref{tab:matching-weight} show that, with larger $\lambda_{m}$, the model achieves higher KL divergence, MI, and reconstruction metrics. Our interpretation is that by pushing the stochastic decoding signals closer to the deterministic ones, we get latent codes with richer text information. We leave the analysis of $\lambda_{m}=0.0$ in Section~\ref{subsec:role-matching}.

The second block in Table~\ref{tab:matching-weight} shows the role of $\lambda_{r}$, which we interpret as follows. When $\lambda_{r}$ is too small (e.g., $0.5$), the learned parameterizations are still inadequate for a smooth transition map; when $\lambda_{r}$ is too large (e.g., $5.0$), it distracts the optimization too far away from the original objective (i.e., $\mathcal{L}_{\itm{rec}} + \mathcal{L}_{\itm{reg}}$). Note that $\lambda_{r}=0.0$ is equivalent to removing the coupled reconstruction loss $\mathcal{L}_{\itm{rec}}^{c}$ in Eq.~(\ref{eq:Coupled-VAE})).

\subsection{The Heterogeneous Effect of Signal Matching on Probability Estimation}
\label{subsec:role-matching}
\begin{table}[t]
    \centering
    \small
    \begin{tabular}{@{}l@{}cccc@{}}
        \toprule
        \multirow{2}*{} & \multicolumn{2}{c}{PTB} & \multicolumn{2}{c}{Yelp}  \\
        \cmidrule(r){2-3} \cmidrule(l){4-5}
        ~                                       & NLL           & PPL           & NLL           & PPL  \\
        \midrule
        Coupled-VAE*                             & 103.1         & 110.5         & 191.2         & 51.6   \\
        Coupled-VAE ($\lambda_{m}$=$0$)            & \bf 103.1     & \bf 110.3     & \bf 190.7     & \bf 51.1 \\
        \midrule
        Coupled-VAE-NF*                          & \bf 102.6     &  \bf 108.1    & \bf 191.8     & \bf 52.2 \\
        Coupled-VAE-NF ($\lambda_{m}$=$0$)         & 102.8         & 109.1         & 192.7         & 53.2 \\
        \midrule
        Coupled-vMF-VAE*                         & \bf 103.0     & \bf 110.1     & \bf 191.2     & \bf 51.6 \\
        Coupled-vMF-VAE ($\lambda_{m}$=$0$) \   &  104.4        &  117.1        & 193.5         & 54.1 \\
        \bottomrule
    \end{tabular}
    \caption{\label{tab:ablation}  The effect of signal matching on probability estimation. * Reported in Table~\ref{tab:lm-results}.}
\end{table}

In Section~\ref{subsec:hyperparameter-analysis} we observe richer latent space (i.e., larger MI and BLEU scores) with larger $\lambda_{m}$. However, a richer latent space does not guarantee a better probability estimation result. Thus, in this part, we delve deeper into whether the decoder signal matching mechanism helps improve probability estimation. We study three models of different posterior families (i.e., Coupled-VAE, Coupled-VAE-NF, and Coupled-vMF-VAE). Results are shown in Table~\ref{tab:ablation}, where we do not report the KL, MI, and BLEU scores because they have been shown to be improved with larger $\lambda_{m}$ in Table~\ref{tab:matching-weight}. We observe that the effects of signal matching on probability estimation vary in different posterior families.

\subsection{Is the Incompatibility Mitigated?}
\label{subsec:analysis-gradients}
\begin{table*}[t]
\centering
\small
\begin{tabular}{@{}l|l|ccccc@{}}
\toprule
\multicolumn{2}{c}{}      & $\left\| \partial \mathcal{L}_{\itm{rec}} / \partial \bm{e} \right\|_{2}$ & $\left\| \partial \mathcal{L}_{\itm{rec}}^{c} / \partial \bm{e} \right\|_{2}$ & $\left\| (\partial \mathcal{L}_{\itm{rec}} + \mathcal{L}_{\itm{rec}}^{c}) / \partial \bm{e} \right\|_{2}$ & $\left\| \partial \mathcal{L}_{\itm{reg}} / \partial \bm{e} \right\|_{2}$ & $\left\| \partial \bm{h} / \partial \bm{e} \right\|_{F} / \left\| \bm{h} \right\|_{2}$ \\
\midrule
\multirow{4}*{PTB}  & DAE               & 1719.8& -         & -         & -     & 3.14 \\
\cmidrule(){2-7}
~                   & VAE               & 112.5 & -         & -         & 19.4  & 2.05 \\
~                   & Coupled-VAE        & \bf 148.5 & 2109.6    &  2320.2   & \bf 27.7  & \bf 2.12 \\
\midrule
\multirow{4}*{Yelp} & DAE               & 2443.6& -         & -         & -     & 2.55 \\
\cmidrule(){2-7}
~                   & VAE               & 59.7  & -         & -         & 18.8  & 1.62 \\
~                   & Coupled-VAE        & \bf 84.8  & 3640.8    & 3764.7    & \bf 25.0  & \bf 2.25 \\
\midrule
\multirow{4}*{Yahoo}& DAE               & 4104.6& -         & -         & -     & 3.39 \\
\cmidrule(){2-7}
~                   & VAE               & 257.9 & -         & -         & 52.8  & 2.92 \\
~                   & Coupled-VAE       & \bf 335.3 & 5105.0    & 5615.0    & \bf 65.0  & \bf 3.91 \\
\bottomrule
\end{tabular}
\caption{\label{tab:gradient-results} Gradient norms defined in Section~\ref{subsec:tracking} on each test set. $\lambda_{m}=0.1$.}
\end{table*}
We study the three gradient norms defined in Section~\ref{sec:analysis} on the test sets, displayed in Table~\ref{tab:gradient-results} (for Coupled-VAE, $\lambda_{m}=0.1$). Notably, $\left\| \partial \mathcal{L}_{\itm{rec}}^{c} / \partial \bm{e} \right\|_{2}$ in Coupled-VAE is even larger than $\left\| \partial \mathcal{L}_{\itm{rec}} / \partial \bm{e} \right\|_{2}$ in DAE. It has two indications. First, the encoder indeed encodes rich information of the text. Second, compared with DAE, Coupled-VAE better generalizes to the test sets, which we conjecture is due to the regularization on the posterior. Coupled-VAE also has a larger $\left\| \partial \mathcal{L}_{\itm{reg}} / \partial \bm{e} \right\|_{2}$ compared with VAE, which based on the argument in Section~\ref{subsec:tracking} indicates that, in Coupled-VAE, the posterior of each instance is not similar to the prior. We also observe larger $\left\| \partial \bm{h} / \partial \bm{e} \right\|_{F} / \left\| \bm{h} \right\|_{2}$ in Coupled-VAE, which indicates a better transition map between the two parameterizations in Coupled-VAE than in VAE.

To show how Coupled-VAE ameliorates the training dynamics, we also track the gradient norms of Coupled-VAE ($\lambda_{m}=10.0$ for a clearer comparison), plotted along with VAE and DAE in Figure~\ref{fig:gradient-norms-curve}. The curve for Coupled-VAE in Figure~\ref{subfig:curve-derecde} stands for $\left\| \partial (\mathcal{L}_{\itm{rec}} + \mathcal{L}_{\itm{rec}}^{c}) / \partial \bm{e} \right\|_{2}$. We observe that Coupled-VAE receives constantly increasing backpropagated gradients from the reconstruction. In contrast to VAE, the $\left\| \partial \mathcal{L}_{\itm{reg}} / \partial \bm{e} \right\|_{2}$ in Coupled-VAE does not decrease significantly as the KL weight increases. The decrease of $\left\| \partial \bm{h} / \partial \bm{e} \right\|_{F} / \left\| \bm{h} \right\|_{2}$, which VAE suffers from, is not observed in Coupled-VAE. Plots on more datasets are in Appendix~\ref{app:tracking-norms}.

\subsection{Sample Diversity}
\label{subsec:diversity-sample}
We evaluate the diversity of the samples from the prior distribution. We sample 3200 texts from the prior distribution and report the Dist-1 and Dist-2 metrics \cite{LiGBGD16}, which are the ratios of distinct unigrams and bigrams over all generated unigrams and bigrams. Distinct-1 and Distinct-2 in Table~\ref{tab:sampled-outputs} show that texts sampled from Coupled-VAE ($\lambda_{m}=10.0$) are more diverse than those from VAE. Given limited space, we put several samples in Appendix~\ref{app:diversity-sample} for qualitative analysis.

\begin{table}[t]
\centering
\small
\begin{tabular}{@{}l@{}cccccc@{}}
\toprule
\multirow{2}*{} & \multicolumn{2}{c}{PTB} & \multicolumn{2}{c}{Yelp} & \multicolumn{2}{c}{Yahoo}\\
\cmidrule(lr){2-3} \cmidrule(lr){4-5} \cmidrule(l){6-7}
~                                       & D-1       & D-2       & D-1       & D-2       & D-1       & D-2 \\
\midrule
VAE                                     & 4.61      & 16.36     & 0.62      & 2.48      & 0.44      & 2.11 \\
Coupled-VAE \ \ \                       & \bf 5.51  & \bf 24.46 & \bf 1.15  & \bf 5.93  & \bf 0.75  & \bf 3.97 \\
\bottomrule
\end{tabular}
\caption{\label{tab:sampled-outputs} Diversity of samples from the prior distribution. \textit{D-} stands for \textit{Distinct-}, normalized to $[0, 100]$.}
\end{table}

\subsection{Interpolation}
\label{subsec:interpolation}
A property of VAE is to match the interpolation in the latent space with the smooth transition in the data space \cite{BowmanVVDJB16}. In Table~\ref{tab:interpolation}, we show the interpolation of VAE and Coupled-VAE on PTB. It shows that compared with VAE, Coupled-VAE has smoother transitions of subjects (\textit{both sides} $\rightarrow$ \textit{it}) and verbs (\textit{are expected} $\rightarrow$ \textit{have been} $\rightarrow$ \textit{has been} $\rightarrow$ \textit{has}), indicating that the linguistic information is more smoothly encoded in the latent space of Coupled-VAE.
\begin{table*}[t]
\begin{center}
\small
\begin{tabular}{@{}ll@{}}
\toprule
\multicolumn{1}{c}{VAE} & \multicolumn{1}{c}{Coupled-VAE ($\lambda_{m}=10.0$)} \\
\midrule
\multicolumn{2}{c}{Text A (sampled from PTB): now those routes are n't expected to begin until jan} \\
\midrule
\multicolumn{1}{@{}p{7.8cm}@{}}{they are n't expected to be completed} & \multicolumn{1}{@{}p{7.8cm}@{}}{both sides are expected to be delivered at their contract} \\
\multicolumn{1}{@{}p{7.8cm}@{}}{the new york stock exchange is scheduled to resume today} & \multicolumn{1}{@{}p{7.8cm}@{}}{both sides are expected to be delivered at least} \\
\multicolumn{1}{@{}p{7.8cm}@{}}{the new york stock exchange is scheduled to resume} & \multicolumn{1}{@{}p{7.8cm}@{}}{both sides have been able to produce up with the current level} \\
\multicolumn{1}{@{}p{7.8cm}@{}}{it is n't clear that it will be sold through its own account} & \multicolumn{1}{@{}p{7.8cm}@{}}{it also has been used for comment} \\
\multicolumn{1}{@{}p{7.8cm}@{}}{it is n't a major source of credit} & \multicolumn{1}{@{}p{7.8cm}@{}}{it also has been working for the first time} \\
\multicolumn{1}{@{}p{7.8cm}@{}}{it also has a major chunk of its assets} & \multicolumn{1}{@{}p{7.8cm}@{}}{it also has a new drug for two years} \\
\multicolumn{1}{@{}p{7.8cm}@{}}{it also has a major pharmaceutical company} & \multicolumn{1}{@{}p{7.8cm}@{}}{it also has a \$ N million defense initiative} \\
\midrule
\multicolumn{2}{c}{Text B (sampled from PTB): it also has a \_unk\_ facility in california}  \\
\bottomrule
\end{tabular}
\caption{\label{tab:interpolation} Latent space interpolation.}
\end{center}
\end{table*}

\subsection{Generalization to Conditional Language Modeling: Coupled-CVAE}
\label{subsec:cvae}
\begin{figure}[t]
\centering
\includegraphics[width=\linewidth]{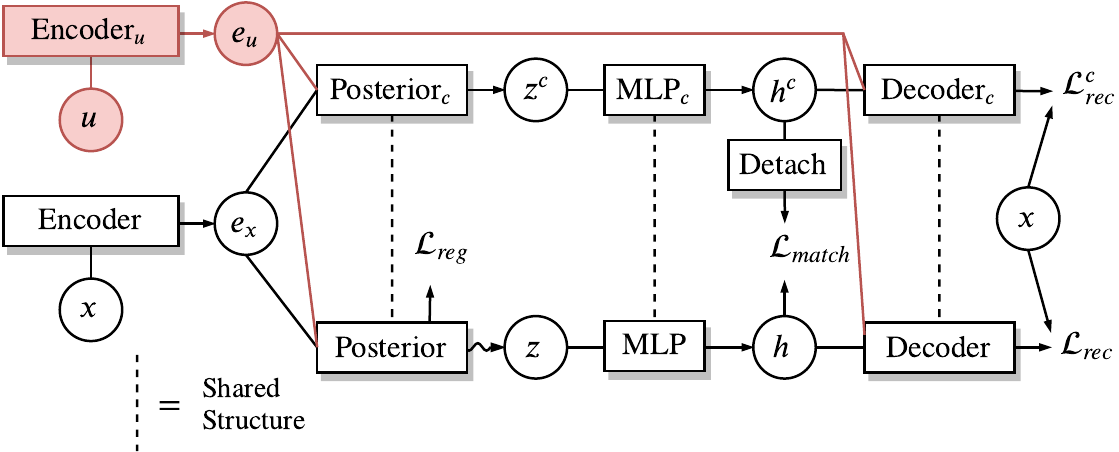}
\caption{A graphical overview of the generalization to Coupled-\textbf{C}VAE. $u$ is the condition, encoded as $\bm{e}_{u}$. The difference from Coupled-VAE is shown in red.}
\label{fig:Coupled-cvae}
\end{figure}
\begin{table}[t]
    \centering
    \small
    \begin{tabular}{@{}l@{}l@{}rrr@{}}
        \toprule
        ~                                       & NLL (KL) \ \ \ & \ PPL       & D-1       & D-2   \\
        \midrule 
        GRU Encoder-Decoder*                    & \bf 53.9 (-)      & \bf 41.6  & 0.33      & 0.80 \\
        \midrule
        CVAE                                    & 54.0 (3.8)        & 41.8      & 0.61      & 2.60 \\
        Coupled-CVAE  ($\lambda_{m}$=$0.1$)  \ \   & 54.1 (4.6)        & 42.2      & 0.71      & 3.18 \\
        Coupled-CVAE  ($\lambda_{m}$=$0.5$)  \ \   & 54.2 (5.3)        & 42.5      & 0.78      & 3.63 \\
        Coupled-CVAE  ($\lambda_{m}$=$1.0$)  \ \   & 54.3 (6.1)        & 42.7      & 0.86      & 4.10 \\
        Coupled-CVAE  ($\lambda_{m}$=$2.0$)  \ \  & 54.6 (7.8)        & 43.6      & \bf 0.99  & \bf 5.16 \\
        \bottomrule
    \end{tabular}
    \caption{\label{tab:cvae-results} Dialogue generation. D-1 and D-2 are normalized to $[0, 100]$. *The exact NLL is reported.}
\end{table}
To generalize our approach to conditional language modeling, we propose Coupled-\textbf{C}VAE. A graphical overview is displayed in Figure~\ref{fig:Coupled-cvae}. Specifically, the (coupled) posterior network and the (coupled) decoder are additionally conditioned. The objective of Coupled-CVAE is identical to Eq.~(\ref{eq:Coupled-VAE}).

We compare Couple-CVAE with GRU encoder-decoder \cite{ChoMGBBSB14} and CVAE \cite{ZhaoZE17} for dialogue generation. We use the Switchboard dataset \cite{Godfrey93}, whose training/validation/test splits are 203K/5K/5K, and the vocabulary size is 13K. For probability estimation, we report the NLL, KL, and PPL based on the gold responses. Since the key motivation of using CVAE in \citet{ZhaoZE17} is the diversity of responses, we sample one response for each post and report the Distinct-1 and Distinct-2 metrics over all test samples. Please find more details of this part in Appendix~\ref{app:coupled-cvae}.

Table~\ref{tab:cvae-results} shows that Coupled-CVAE greatly increases the diversity of dialogue modeling, while it only slightly harms the probability estimation capability. It indicates that Coupled-CVAE better captures the one-to-many nature of conversations than CVAE and GRU encoder-decoder. We also observe that the diversity is improved with increasing $\lambda_{m}$, which shows that $\lambda_{m}$ can control diversity via specifying the richness of the latent space.

\section{Relation to Related Work}
\citet{BowmanVVDJB16} identify the posterior collapse problem of text VAE and propose KL annealing and word drop to handle the problem. \citet{ZhaoZE17} propose the bag-of-words loss to mitigate this issue. Later work on this problem focuses on less powerful decoders \cite{YangHSB17,SemeniutaSB17}, modified regularization objective \cite{HigginsMPBGBML17,BahuleyanMZV19,WangW19}, alternative posterior families \cite{RezendeM15,XuD18,DavidsonFCKT18,XiaoDirichlet18}, richer prior distributions \cite{TomczakW18}, improved optimization \cite{HeSNB19} or KL annealing strategy \cite{FuLLGCC19}, the use of skip connections \cite{DiengKRB19}, hierarchical or autoregressive posterior distributions \cite{ParkCK18,DuLHXBW18}, and narrowing the amortization gap \cite{HjelmSCJCC16,KimWMSR18,MarinoYM18}. We provide the encoder-decoder incompatibility as a new perspective on the posterior collapse problem. Empirically, our approach can be combined with the above ones to alleviate the problem further.

A model to be noted is $\beta$-VAE \cite{HigginsMPBGBML17}, in which the reconstruction and regularization are modeled as a hyperparameterized trade-off, i.e., the improvement of one term compromises the other. Different from $\beta$-VAE, we adopt the idea of multi-task learning, i.e., the \textit{coupled reconstruction} task helps improve the encoder chart map and the \textit{signal matching} task helps improve the decoder chart map. Both our analysis in Section~\ref{subsec:incompatibility} and the empirical results show that the modeling of posterior distribution can be improved (but not necessarily compromised) with the additional tasks.

\citet{Ghosh19deterministic} propose to substitute stochasticity with explicit and implicit regularizations, which is easier to train and empirically improves the quality of generated outputs. Different from their work, we still strictly follow the generative nature (i.e., data density estimation) of VAE, and the deterministic network in our approach serves as an auxiliary to aid the optimization.

Encoder pretraining \cite{LiHNBY19} initializes the text encoder and the posterior network with an autoencoding objective. \citet{LiHNBY19} shows that encoder pretraining itself does not improve the performance of VAE, which indicates that initialization is not strong enough as an inductive bias to learn a meaningful latent space. 

Given the discrete nature of text data, we highlight the two-level representation learning for text modeling: 1) the encoder and decoder parameterizations via autoencoding and 2) a transition map between the parameterizations. Notably, the transition map has large freedom. In our case, the transition map decides the amount and type of information encoded in the variational posterior, and there are other possible instances of the transition map, e.g., flow-based models \cite{DinhKB14}. 

\section{Conclusions}
In this paper, we observe the encode-decoder incompatibility of VAE for text modeling. We bridge the incompatibility and the posterior collapse problem by viewing the encoder and the decoder as two inadequately learned chart maps from the data manifold to the parameterizations, and the posterior network as a part of the transition map between them. We couple the VAE model with a deterministic network and improve the parameterizations via encoder weight sharing and decoder signal matching. Our approach is model-agnostic and can be applied to a wide range of models in the VAE family. Experiments on benchmark datasets, i.e., PTB, Yelp, and Yahoo, show that our approach improves various VAE models in terms of probability estimation and the richness of the latent space. We also generalize Coupled-VAE to conditional language modeling and propose Coupled-CVAE. Results on Switchboard show that Coupled-CVAE largely improves diversity in dialogue generation.

\section*{Acknowledgments}
We would like to thank the anonymous reviewers for their thorough and helpful comments. 


\clearpage

\bibliography{acl2020}
\bibliographystyle{acl_natbib}

\clearpage 

\appendix
\section*{Appendix}
\section{Notations}
We first introduce the notations used in the following parts. Calligraphic letters (e.g., $\mathcal{Q}_{0}$) denotes continuous distributions, and the corresponding lowercase letters (e.g., $q_{0}$) stands for probability density functions. The probability of the text is represented as $P$.

\section{Deterministic Networks for Different Posterior Families}
\label{app:deterministic-details}
In this part, we detail the forward computation of the deterministic networks for different posterior families, including multivariate Gaussian, Gaussian with normalizing flows, and von Mises–Fisher. 

\subsection{Multivariate Gaussian}
For multivariate Gaussian, we compute the coupled latent code $\bz^{c}$ as
\begin{equation}
\label{eq:zc-gaussian}
    \bz^{c} = \mathbb{E}_{\bz \sim \mathcal{Q}^{c}(\bz|x)}[\bz]
\end{equation}
where $\mathcal{Q}^{c}(\bz|x)$ is the posterior distribution learned by the coupled deterministic network. In effect, $\bz$ is the mean vector predicted by the coupled posterior network $\mathrm{Posterior}^{c}$.

\subsection{Gaussian with Normalizing Flows}
We first review the background and notations of normalizing flows. An initial latent code is first sampled from an initial distribution, i.e., $\bz_{0} \sim \mathcal{Q}_{0}(\bz_{0} | x)$. The normalizing flow is defined as a series of \textit{reversible} transformations $f_{1}, \ldots, f_{K}$, i.e.,
\begin{equation}
\label{eq:flows}
\begin{split}
    \bz_{k} &= f_{k} \circ \cdots \circ f_{1} (\bz_{0})
\end{split}
\end{equation}
where $k = 1, \ldots, K$. 
The evidence lower bound (ELBO) for normalizing flows is derived as
\begin{equation}
\label{eq:nf-elbo}
\begin{split}
    \log P(x) &\geq \mathbb{E}_{\bz_{K} \sim \mathcal{Q}_{K}(\bz_{K}|x)}[\log P(x|\bz_{K})] \\
    &\ \ \ \ \ - \mathrm{KL}[\mathcal{Q}_{K}(\bz_{K}|x) \parallel \mathcal{P}_{K}(\bz_{K})] \\
    &= \mathbb{E}_{\bz_{0} \sim \mathcal{Q}_{0}(\bz_{0}|x)} \Big [\log P(x|\bz_{K}) \\
    &\ \ \ \ \ - \log q_{0}(\bz_{0}|x) + \log p_{K}(\bz_{K}) \\
    &\ \ \ \ \ + \sum_{k=1}^{K}\log |\det \frac{\partial f_{k}}{\partial \bz_{k-1}}|\Big]
\end{split}
\end{equation}
where $\mathcal{P}_{K}(\bz_{K})$ is the prior distribution of the transformed latent variable and the reversibility of the transformations guarantees non-zero determinants. Obviously, the optimization of the ELBO for normalizing flows requires sampling from the initial distribution; thus, we compute the coupled latent code $\bz^{c}$ by transforming the predicted mean vector of the coupled initial distribution, i.e.,
\begin{equation}
\label{eq:zc-normalizing-flows}
    \bz^{c} = f^{c}_{k} \circ \cdots \circ f^{c}_{1} (\mathbb{E}_{\bz_{0} \sim \mathcal{Q}_{0}^{c}(\bz_{0}|x)}[\bz_{0}])
\end{equation}
where $\mathcal{Q}_{0}^{c}(\bz_{0}|x)$ is the coupled initial distribution and $f^{c}_{1}, \ldots, f^{c}_{K}$ are the coupled transformations. Note that all modules in the deterministic network share the structure with those in the stochastic network. We do not use the posterior mean as the coupled latent code for two reasons. First, our interest is to acquire a deterministic representation that guides the stochastic network, but not necessarily the mean vector. Second, the computation of the posterior mean after the transformations is intractable. 

\subsection{Von Mises-Fisher}
The von Mises-Fisher distribution is supported on a $(d-1)$-dimensional sphere in $\mathbb{R}^{d}$ and parameterized by a direction parameter $\bm{\mu} \in \mathbb{R}^{d}$ ($\lVert \bm{\mu} \rVert = 1$) and a concentration parameter $\kappa$, both of which are mapped from the encoded text by the posterior network. The probability density function is
\begin{equation}
\label{eq:vmf-density}
    q(\bz|\bm{\mu}, \kappa) = \frac{\kappa^{d/2 - 1} \cdot \exp(\kappa\bm{\mu}^{\mathrm{T}}\bz)}{(2\pi)^{d/2}I_{d/2 - 1}(\kappa)}
\end{equation}
where $I_v$ is the modified Bessel function of the first kind at order $v$. We use the direction parameter $\bm{\mu}$ as the coupled latent code $\bz^{c}$. Note that we do not use the posterior mean as the coupled latent code for two reasons. First, similar to normalizing flows, our interest is a deterministic representation rather than the mean vector. Second, the posterior mean of von Mises-Fisher \textit{never} lies on the support of the distribution, which is suboptimal to guide the stochastic network.

\section{Details of the Experimental Setup}
\label{app:setup}
The dimension of latent vectors is 32. The dimension of word embeddings is 200. The encoder and the decoder are one-layer GRUs with the hidden state size of 128 for PTB and 256 for Yelp and Yahoo. For optimization, we use Adam \cite{KingmaB14} with a learning rate of $10^{-3}$ and $\beta_{1} = 0.9$, $\beta_{1} = 0.999$. The decoding signal is viewed as the first word embedding and also concatenated to the word embedding in each decoding step. After 30K steps, the learning rate is decayed by half each 2K steps. Dropout \cite{SrivastavaHKSS14} rate is $0.2$. KL-annealing \cite{BowmanVVDJB16} is applied from step 2K to 42K (on Yelp, it is applied from step 1K to 41K for VAE, Coupled-VAE, $\beta$-VAE, and Coupled-$\beta$-VAE; otherwise, the KL divergence becomes very large in the early stage of training). For each 1K steps, we estimate the NLL for validation. 

For normalizing flows (NF), we use planar flows \cite{RezendeM15} with three contiguous transformations. For WAE and WAE-NF, we use Maximum Mean Discrepancy (MMD) \cite{GrettonBRSS12} as the regularization term. An additional KL regularization term with the weight $\beta=0.8$ (also with KL-annealing) is added to WAE and WAE-NF since MMD does not guarantee the convergence of the KL divergence. 

\section{Estimation of Language Modeling Metrics}
\label{app:lm-metrics}
For language modeling, we report negative log-likelihood (NLL), KL divergence, and perplexity. To get more reliable results, we make the estimation of each metric explicit. For each test sample $x$, NLL is estimated by importance sampling, and KL is approximated by its Monte Carlo estimate:
\begin{equation}
\begin{split}
\label{eq:iw-nll-mc-kl}
    \textrm{NLL}_{x} &= - \log P(x) \\
    &\approx - \log (\frac{1}{N} \sum_{i=1}^{N} \frac{p(\bz^{(i)})P(x|\bz^{(i)})}{q(\bz^{(i)}|x)}) \\
    \textrm{KL}_{x} &= \mathrm{KL}[\mathcal{Q}(\bz|x) \parallel \mathcal{P}(\bz)] \\
    & \approx \frac{1}{N} \sum_{i=1}^{N} \log \frac{q(\bz^{(i)}|x)}{p(\bz^{(i)})}
\end{split}
\end{equation}
where $\bz^{(i)} \sim \mathcal{Q}(\bz|x)$ are sampled latent codes and all notations follow
Eq.~(\ref{eq:vae-elbo}) in the main text. 
We report the averaged NLL and KL on all test samples. Perplexity is computed based on the estimated NLL. For validation, the number of samples is $N=10$; for evaluation, the number of samples is $N=100$. 

\begin{figure*}[t]
\centering
    \subfigure[PTB]{
        \includegraphics[width=0.3\linewidth]{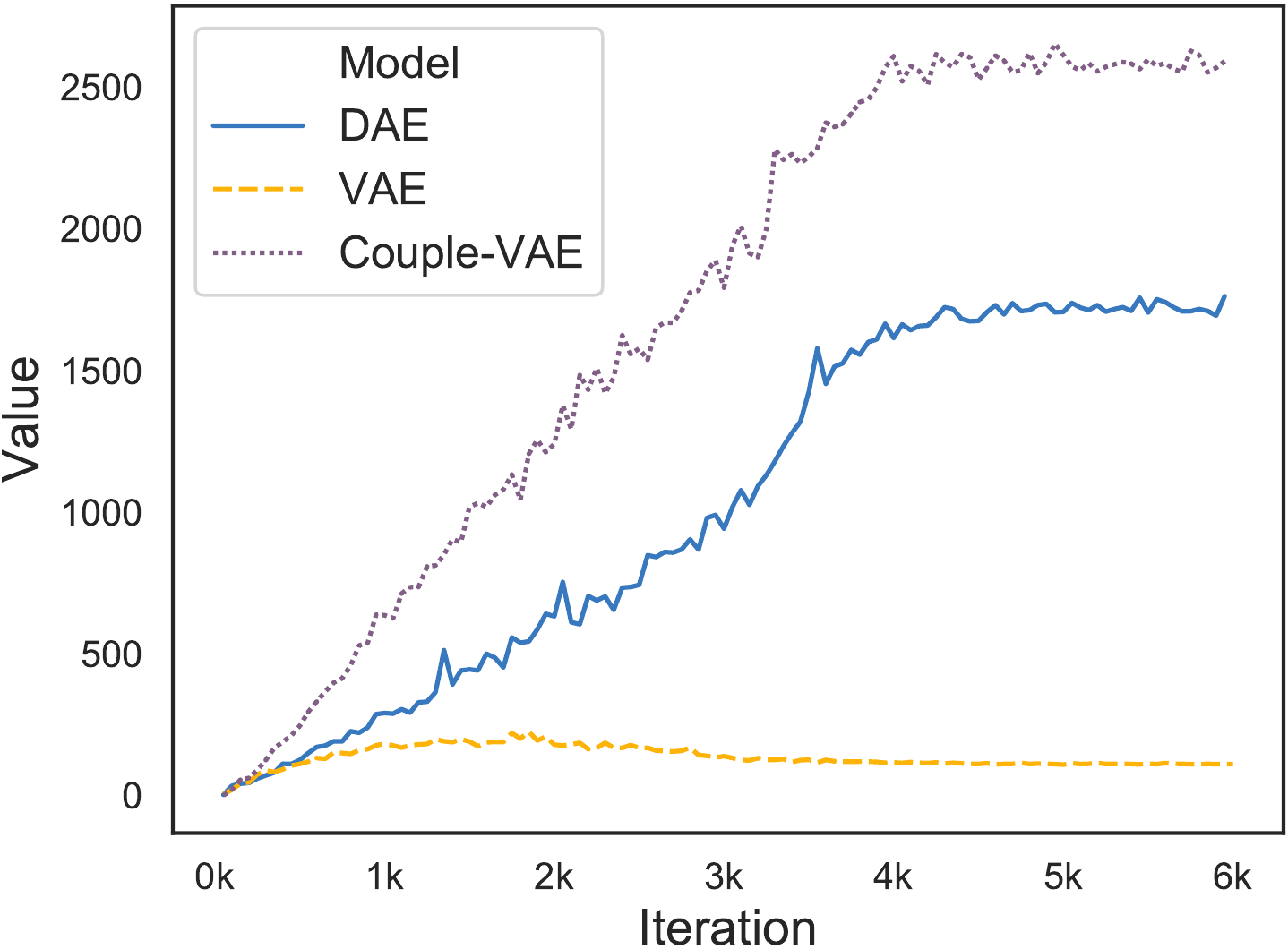}
        \label{subfig:curve-derecde-ptb}
    }
\hspace{0.01\linewidth}
    \subfigure[PTB]{
        \includegraphics[width=0.3\linewidth]{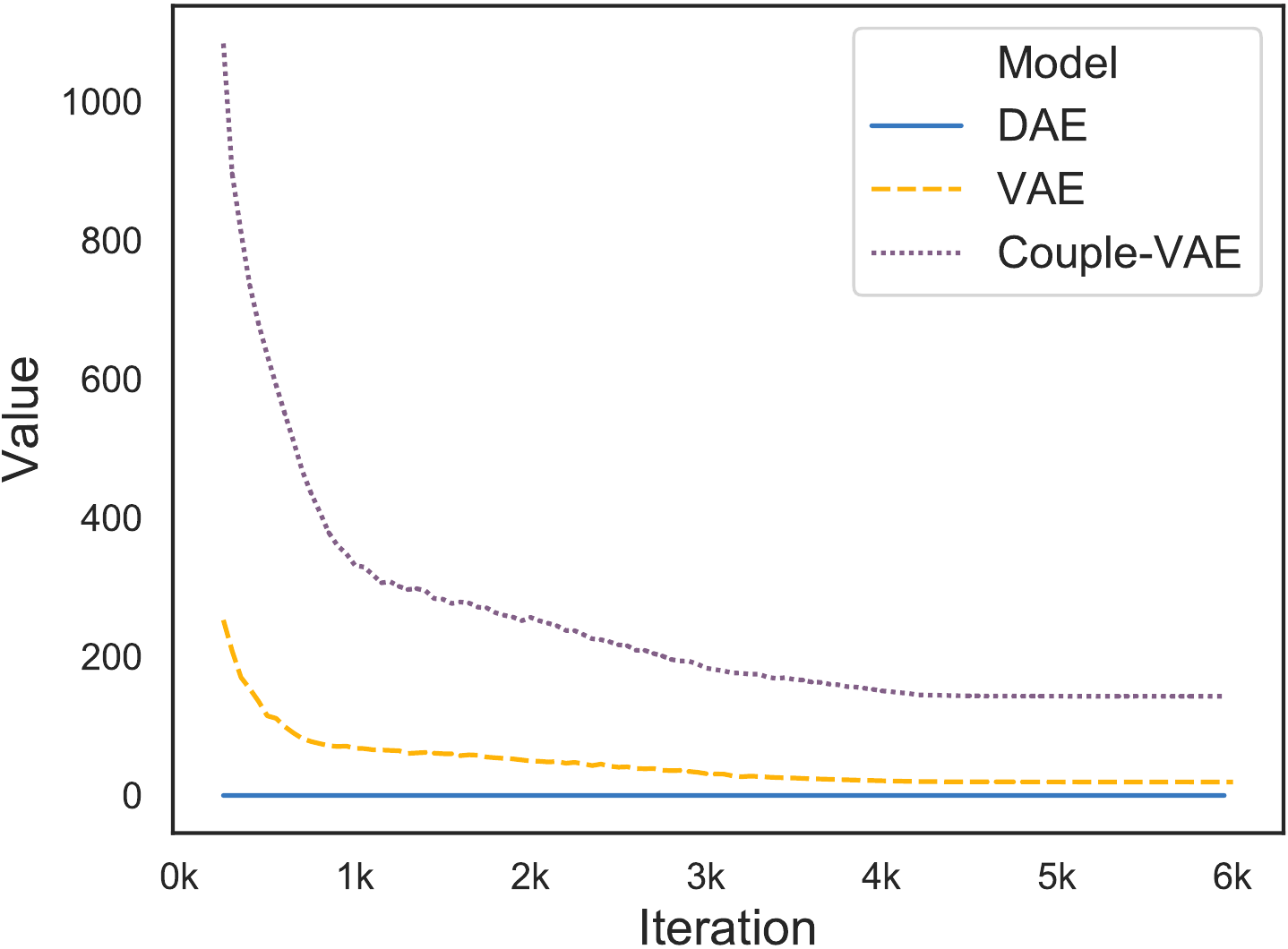}
        \label{subfig:cuurve-dregde-ptb}
    }
\hspace{0.01\linewidth}
    \subfigure[PTB]{
        \includegraphics[width=0.3\linewidth]{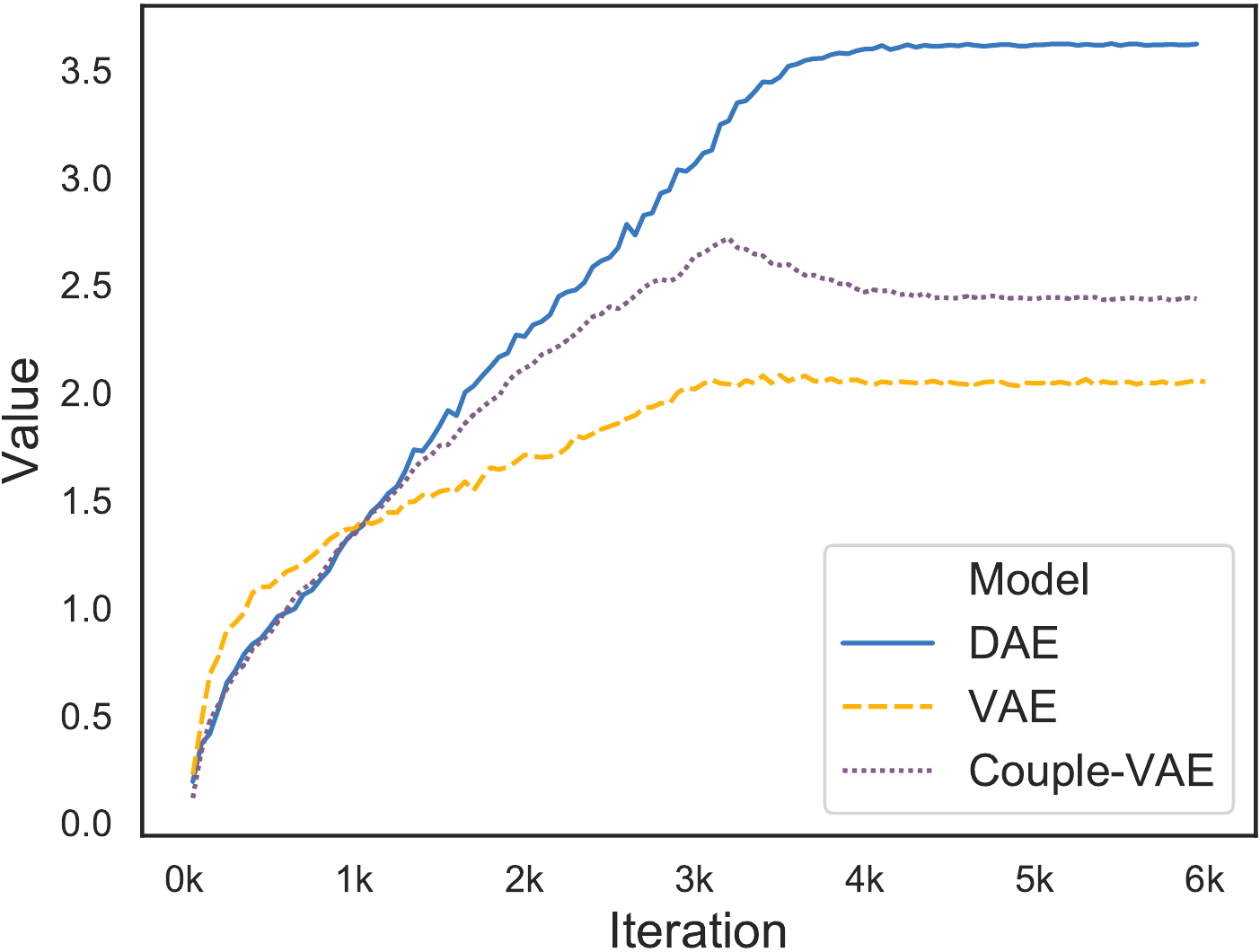}
        \label{subfig:cuurve-dhde-ptb}
    }
    \subfigure[Yelp]{
        \includegraphics[width=0.3\linewidth]{drecde-yelp.pdf}
        \label{subfig:curve-derecde-yelp}
    }
\hspace{0.01\linewidth}
    \subfigure[Yelp]{
        \includegraphics[width=0.3\linewidth]{dregde-yelp.pdf}
        \label{subfig:cuurve-dregde-yelp}
    }
\hspace{0.01\linewidth}
    \subfigure[Yelp]{
        \includegraphics[width=0.3\linewidth]{dhde-yelp.pdf}
        \label{subfig:cuurve-dhde-yelp}
    }
    \subfigure[Yahoo]{
        \includegraphics[width=0.3\linewidth]{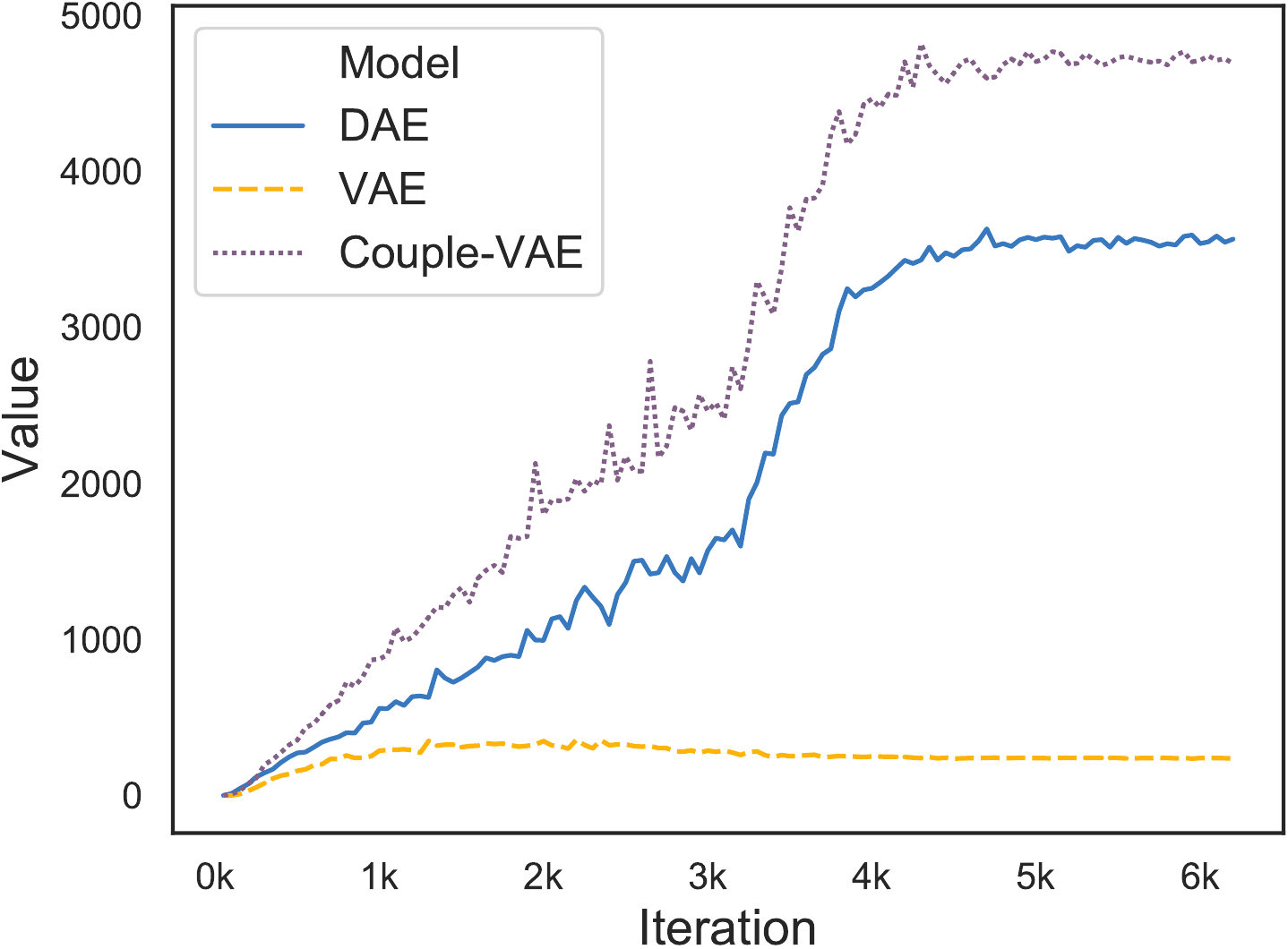}
        \label{subfig:curve-derecde-yahoo}
    }
\hspace{0.01\linewidth}
    \subfigure[Yahoo]{
        \includegraphics[width=0.3\linewidth]{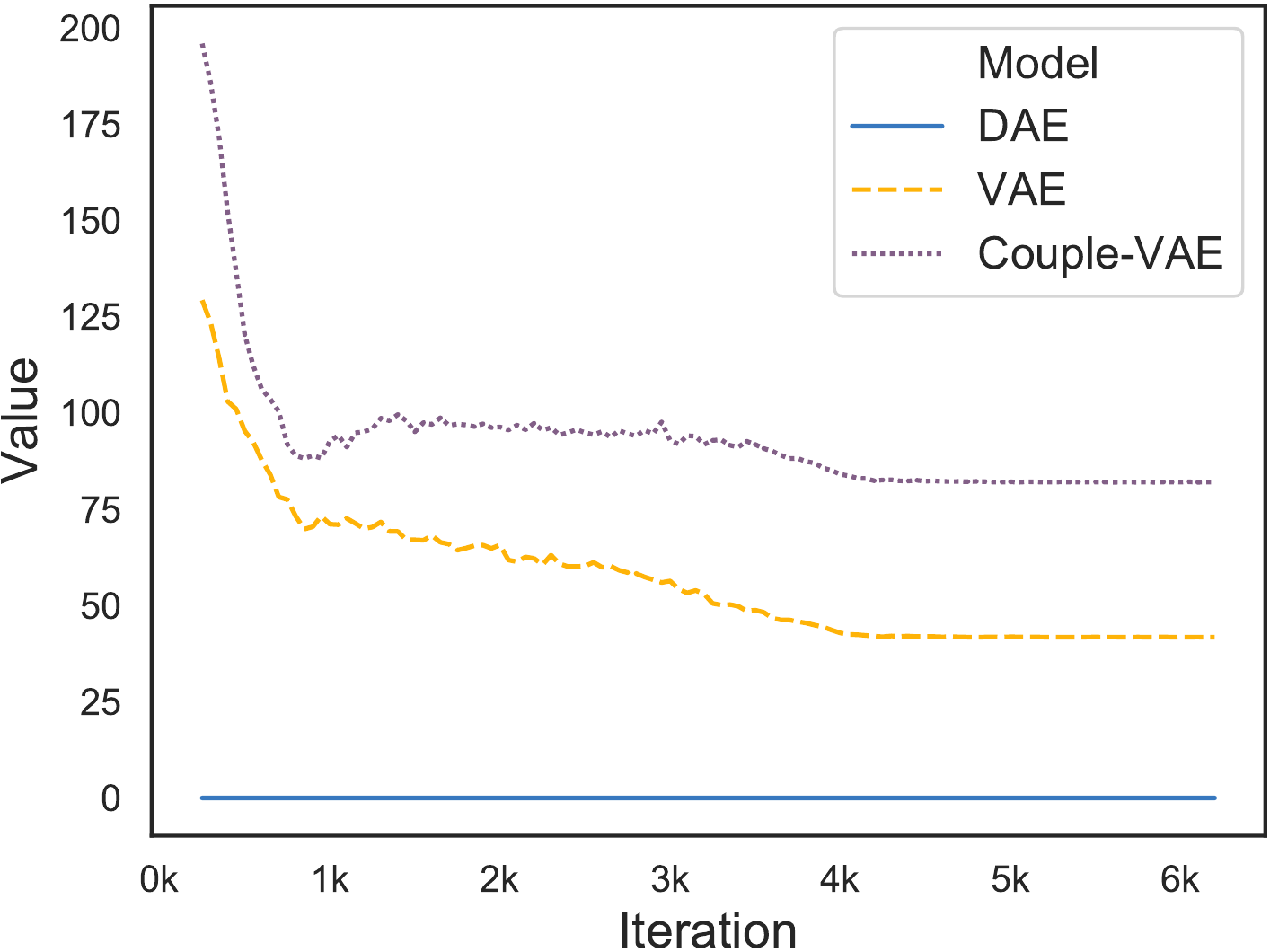}
        \label{subfig:cuurve-dregde-yahoo}
    }
\hspace{0.01\linewidth}
    \subfigure[Yahoo]{
        \includegraphics[width=0.3\linewidth]{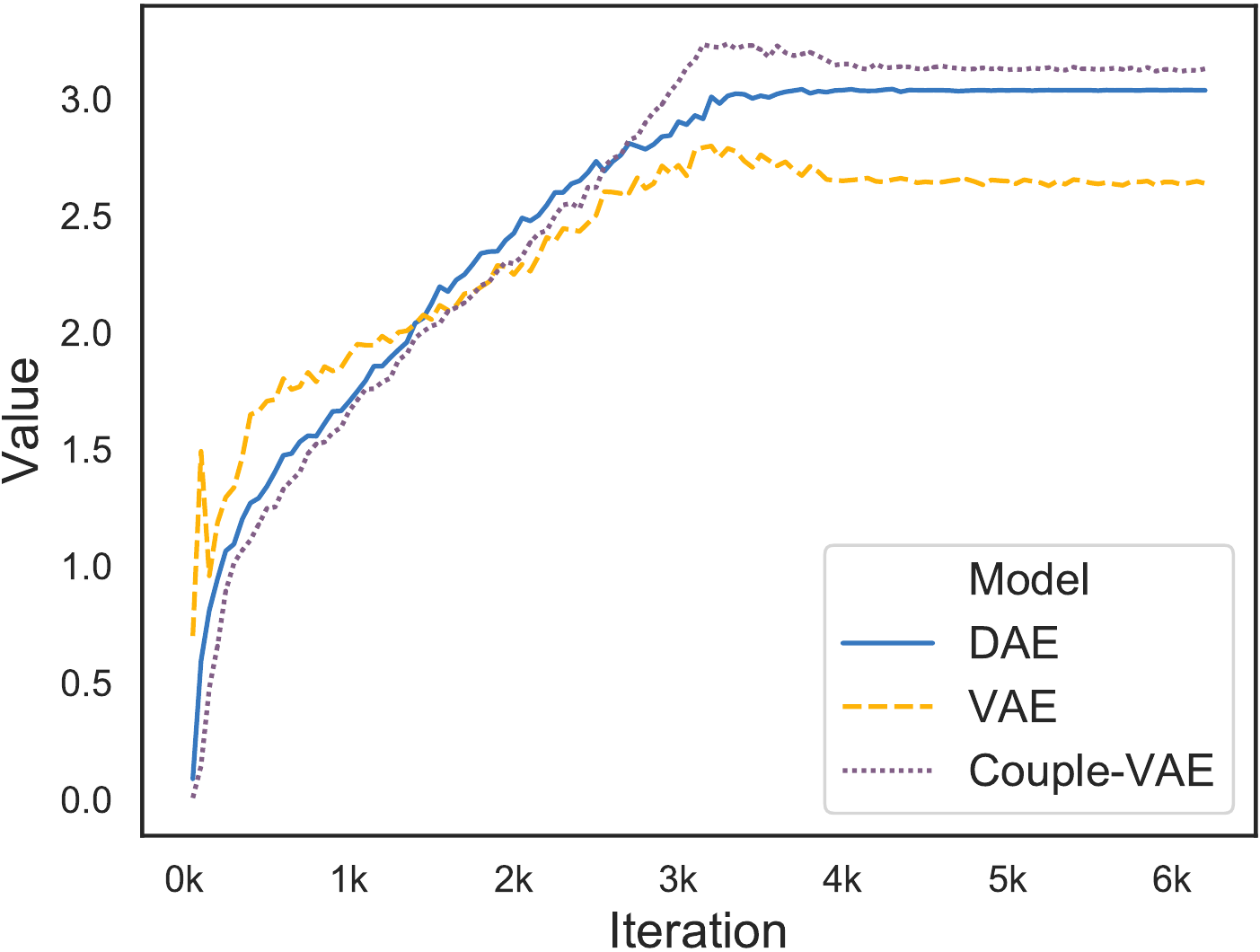}
        \label{subfig:cuurve-dhde-yahoo}
    }    
\caption{\label{fig:gradient-norms-curve-all} Training dynamics of DAE, VAE, and Coupled-VAE ($\lambda_{m} = 10.0$). (a), (d), and (g) are $\left\| \partial \mathcal{L}_{\itm{rec}} / \partial \bm{e} \right\|_{2}$ for DAE and VAE, and $\left\| \partial (\mathcal{L}_{\itm{rec}} + \mathcal{L}_{\itm{rec}}^{c}) / \partial \bm{e} \right\|_{2}$ for Coupled-VAE. (b), (e), (h) denote $\left\| \partial \mathcal{L}_{\itm{reg}} / \partial \bm{e} \right\|_{2}$. (c), (f), (i) stand for $\left\| \partial \bm{h} / \partial \bm{e} \right\|_{F} / \left\| \bm{h} \right\|_{2}$. Best viewed in color (yet the models are distinguished by line markers).}
\end{figure*}

\section{Estimation of Mutual Information and Reconstruction Metrics}
\label{app:info-metrics}
We report the mutual information (MI) between the text $x$ and the latent code $\bz$ under $\mathcal{Q}(\bz|x)$ to investigate how much useful information is encoded. The MI component of each test sample $x$ is approximated by Monte Carlo estimation:
\begin{equation}
\label{eq:mc-mi}
\begin{split}
    \textrm{MI}_{x} &= \mathbb{E}_{\bz \sim \mathcal{Q}(\bz|x)}[\log \frac{q(\bz|x)}{q(\bz)}] \\
    & \approx \frac{1}{N} \sum_{i=1}^{N} (\log q(\bz^{(i)}|x) - \log q(\bz^{(i)}))
\end{split}
\end{equation}
where the aggregated posterior density $q(\bz^{(i)})$ is approximated with its Monte Carlo estimate:
\begin{equation}
\small
\label{eq:aggregate-posterior}
\begin{split}
    q(\bz^{(i)}) &= \mathbb{E}_{x}[q(\bz^{(i)}|x)] \approx \frac{1}{M} \sum_{j=1}^{M}q(\bz^{(i)}|x^{(j)})
\end{split}
\end{equation}
where $x^{(j)}$ are sampled from the test set. For convenience, most previous work uses the texts within each batch as the sampled $x^{(j)}$'s (which are supposed to be sampled from the entire test set). However, this convention results in a biased estimation since the $q(\bz^{(i)}|x^{(i)})$ is computed when $j=i$, i.e., the text itself is always sampled when computing its MI component. We remedy it by skipping the term when $j=i$. The overall $\textrm{MI} = \mathbb{E}_{x}[\textrm{MI}_{x}]$ is then estimated by averaging $\textrm{MI}_{x}$ over all test samples. We set the numbers of samples as $N=100$ and $M=512$.

For reconstruction, we sample ten latent codes from the posterior of each text input and decode them with greedy search. We compute BLEU-1 and BLEU-2 between the reconstruction and the input with the Moses script.

\section{Training Dynamics of Gradient Norms}
\label{app:tracking-norms}
We show the tracked gradient norms on all datasets in Figure~\ref{fig:gradient-norms-curve-all}. The observations are consistent with those discussed in Section~\ref{subsec:analysis-gradients} in the main text.

\section{Diversity and Samples from the Prior Distribution}
\label{app:diversity-sample}
Given the limited space in the main text, we place the comprehensive evaluation of samples from the prior distribution in this part. Table~\ref{tab:sampled-outputs-all} shows the diversity metrics and the first three (thus totally random) samples from each model. Qualitatively, samples from Coupled-VAE is more diverse than those from VAE. The long texts generated from VAE have more redundancies compared with Coupled-VAE. Given that both models have the same latent dimension, it indicates that Coupled-VAE is using the latent codes more efficiently.
\begin{table*}[t]
\centering
\small
\begin{tabular}{@{}lll@{}}
\toprule
VAE (PTB)  & Dist-1 = 0.0461 & Dist-2 = 0.1636 \\
\midrule
\multicolumn{3}{@{}p{13.9cm}@{}}{1. but the market is a bit of the market 's recent slide and the fed is trying to sell investors to buy back and forth between the s\&p N and N} \\
\multicolumn{3}{@{}p{13.9cm}@{}}{2. the company said it will be developed by a joint venture with the u.s.} \\
\multicolumn{3}{@{}p{13.9cm}@{}}{3. the new york stock exchange composite index rose N to N} \\
\midrule
Coupled-VAE ($\lambda_{m}$ = 10.0) (PTB) & Dist-1 = \textbf{0.0551} & Dist-2 = \textbf{0.2446} \\
\midrule
\multicolumn{3}{@{}p{13.9cm}@{}}{1. dd acquisition said it will offer to acquire N shares of lin 's shares to be sold} \\
\multicolumn{3}{@{}p{13.9cm}@{}}{2. but the u.s. would be closed at N p.m. edt in N but that was caused by lower rates} \\ 
\multicolumn{3}{@{}p{13.9cm}@{}}{3. \$ N billion in the stock market was a lot of it to be worth for each of N} \\
\bottomrule
\toprule
VAE (Yelp)  & Dist-1 = 0.0062 & Dist-2 = 0.0248 \\
\midrule
\multicolumn{3}{@{}p{13.9cm}@{}}{1. the food is good , but \red{the food is good} . i had the chicken fried steak with a side of mashed potatoes , and it was a good choice . the fries were good , but \red{the fries were good} . \red{i had the chicken} breast \red{with a side}} \\
\multicolumn{3}{@{}p{13.9cm}@{}}{2. ok , so i was excited to check out this place for a while . i was in the area , and i was n't sure what to expect . i was a little disappointed with the food , but \red{i was n't sure what to expect . i was}} \\
\multicolumn{3}{@{}p{13.9cm}@{}}{3. we went to the biltmore fashion park . we were seated right away , but \red{we were seated right away} . \red{we were seated right away , but we were seated right away . we were seated right away and we were seated right away .} the staff was very} \\
\midrule
Coupled-VAE ($\lambda_{m}=10.0$) (Yelp) & Dist-1 = \textbf{0.0115} & Dist-2 = \textbf{0.0593} \\
\midrule
\multicolumn{3}{@{}p{13.9cm}@{}}{1. i 'm a fan of the `` asian '' restaurants in the valley , and i 'm not sure what to expect , but \red{i 'm not sure what} the fuss is about . the meat is fresh and delicious . i 'm not \red{a fan of} the `` skinny} \\
\multicolumn{3}{@{}p{13.9cm}@{}}{2. i 'm not a fan of the fox restaurants in phoenix , but i have to say that the service is always a great experience . the atmosphere is a little dated and there is a great view of the mountains .} \\ 
\multicolumn{3}{@{}p{13.9cm}@{}}{3. i have been here twice , and the food was good , but the service was good , but \red{the food was good} . i had a great time , but \red{the service was} great . \red{the food was} a bit pricey , but \red{the service was} a bit slow} \\
\bottomrule
\toprule
VAE (Yahoo)  & Dist-1 = 0.0044 & Dist-2 = 0.0211 \\
\midrule
\multicolumn{3}{@{}p{13.9cm}@{}}{1. what is the difference between the two and the \_UNK ? i am not sure what you mean , but \red{i 'm not sure what you mean . i 'm not sure what you mean , but i 'm not sure what you mean .} the answer is : 1 . the first person is the first person to be \red{the first person to be the first person to be the first person .} 2 . \red{the first person is the first person to be the first person to be the first person . the first thing is that the person who is the best person is to be a person ,} and \red{the person who is the} best person to be born . \red{the person who is not the best person is to be a person , and the person who is not the best person to be born .}} \\
\multicolumn{3}{@{}p{13.9cm}@{}}{2. what do you think of the song `` \_UNK '' ? i 'm not sure what you 're talking about . \red{i 'm not sure what you 're talking about . i 'm not sure what you 're talking about . i 'm not sure what you 're talking about . i 'm not sure what you 're talking about . i 'm not sure what you 're talking about . i 'm not sure what you 're talking about .}} \\
\multicolumn{3}{@{}p{13.9cm}@{}}{3. what is the name of the song ? i heard that the song was a song called `` \_UNK '' . it was \red{a song called `` \_UNK '' . it was a song called `` \_UNK '' . it was a song called `` \_UNK '' . it was a song called `` \_UNK '' . it was a song called `` \_UNK '' . it was a song called `` \_UNK '' , `` \_UNK '' , `` \_UNK '' , `` \_UNK '' , `` \_UNK '' , `` \_UNK '' , `` \_UNK '' , `` \_UNK '' , `` \_UNK '' , `` \_UNK '' , `` \_UNK '' , `` \_UNK '' , `` \_UNK '' , `` \_UNK '' , `` \_UNK '' , `` \_UNK '' , `` \_UNK '' , `` \_UNK '' , `` \_UNK '' , `` \_UNK '' , `` \_UNK '' , `` \_UNK '' , `` \_UNK '' , `` \_UNK '' , `` \_UNK '' , `` \_UNK ''}
} \\
\midrule
Coupled-VAE ($\lambda_{m}=10.0$) (Yahoo) & Dist-1 = \textbf{0.0075} & Dist-2 = \textbf{0.0397} \\
\midrule
\multicolumn{3}{@{}p{13.9cm}@{}}{1. if you are looking for a good wrestler , what do you think about the future ? i am not sure what i mean . i have been watching the ufc for 3 months . \red{i have been watching the ufc} and i have to be able to see what happens .} \\
\multicolumn{3}{@{}p{13.9cm}@{}}{2. is it true that the war is not a hoax ? it is a myth that the \_UNK of the war is not a war , but it is not possible to be able to see the war . the \_UNK \red{is not a war , but it 's} not a crime .} \\ 
\multicolumn{3}{@{}p{13.9cm}@{}}{3. how do i get a \_UNK on ebay ? ebay is free and they are free !} \\
\bottomrule
\end{tabular}
\caption{\label{tab:sampled-outputs-all} Diversity metrics and the first three samples from each model. Redundancies (pieces of text that appeared before) are shown in red.}
\end{table*}
\section{Interpolation}
\label{app:interpolation}
A property of VAE is to match the interpolation in the latent space with the smooth transition in the text space \cite{BowmanVVDJB16}. In Table~\ref{tab:interpolation}, we show the interpolation of VAE and Coupled-VAE on PTB. It shows that compared with VAE, Coupled-VAE has smoother transitions of subjects (\textit{both sides} $\rightarrow$ \textit{it}) and verbs (\textit{are expected} $\rightarrow$ \textit{have been} $\rightarrow$ \textit{has been} $\rightarrow$ \textit{has}), indicating that the information about subjects and verbs is more smoothly encoded in the latent space of Coupled-VAE.

\section{Generalization to Conditional Generation: Coupled-CVAE}
\label{app:coupled-cvae}
To generalize our approach to conditional generation, we focus on whether it can improve the CVAE model \cite{ZhaoZE17} for dialogue generation. To this end, we propose the Coupled-CVAE model. 
\subsection{CVAE}
CVAE adopts a two-step view of diverse dialogue generation. Let $x$ be the response and $y$ be the post (or the context). CVAE first samples the latent code $\bz$ from the prior distribution $\mathcal{P}(\bz|y)$ and then samples the response from the decoder $P(x|\bz, y;\theta)$. Given the post $y$, the marginal distribution of the response $x$ is 
\begin{equation}
\label{eq:marginal-cvae}
    P(x|y; \theta) = \mathbb{E}_{\bz \sim \mathcal{P}(\bz|y)}[P(x|\bz,y;\theta)]
\end{equation}
Similar to VAE, the exact marginalization is intractable, and we derive the evidence lower bound (ELBO) of CVAE as
\begin{equation}
\label{eq:cvae-elbo}
\begin{split}
    \log P(x|y;\theta) &\geq \mathbb{E}_{\bz \sim \mathcal{Q}(\bz|x, y;\phi)}[\log P(x|\bz, y;\theta)] \\
    &\ \ \ \ \ - \mathrm{KL}[\mathcal{Q}(\bz|x, y; \phi) \parallel \mathcal{P}(\bz|y)]
\end{split}
\end{equation}
During training, the response and the post are encoded as $\bm{e}_{x}$ and  $\bm{e}_{y}$, respectively. The two vectors are concatenated and transformed into the posterior via the posterior network. A latent code is then sampled and mapped to a higher-dimensional $\bm{h}$. The decoding signal in CVAE is computed by $\bm{h}$ and $\bm{e}_{y}$ and utilized to infer the response. Similar to VAE, the objective of CVAE can also be viewed as a reconstruction loss and a regularization term in Eq.~(\ref{eq:cvae-elbo}). 

\subsection{Coupled-CVAE}
As observed in \citet{ZhaoZE17}, the CVAE model also suffers from the posterior collapse problem. We generalize our approach to the conditional setting and arrive at Coupled-CVAE. A graphical overview is displayed in Figure~\ref{fig:Coupled-cvae}. The difference from Coupled-VAE is shown in red. Specifically, the (coupled) posterior network and the (coupled) decoder are additionally conditioned on the post representation. The objective of Coupled-CVAE is identical to Eq.~(\ref{eq:Coupled-VAE}) in the main text.

The coupled reconstruction loss $\mathcal{L}_{\itm{rec}}^{c}$ in Coupled-CVAE has two functions. First, it improves the encoded response $\bm{e}_{x}$, which is similar to Coupled-VAE. Second, it encourages $\bm{h}^{c}$ to encode more response information rather than the post information, which collaborates with $\mathcal{L}_{\itm{match}}$ to improve the parameterization $\bm{h}$.

\subsection{Dataset}
We use the Switchboard dataset \cite{Godfrey93}. We split the dialogues into single-turn post-response pairs, and the number of pairs in the training/validation/test split is 203K/5K/5K. The vocabulary size is 13K. 

\subsection{Evaluation} 
For probability estimation, we report the NLL, KL, and PPL based on the gold responses. NLL, KL, and PPL are as computed in Appendix~\ref{app:lm-metrics} except for the additional condition on the post. Since the key motivation of using CVAE in \citet{ZhaoZE17} is the response diversity, we sample one response for each post and report the Distinct-1 and Distinct-2 metrics over all test samples.

\subsection{Experimental Setup}
We compare our Coupled-CVAE model with two baselines: GRU encoder-decoder \cite{ChoMGBBSB14} and CVAE \cite{ZhaoZE17}. The detailed setup follows that of the PTB dataset in Appendix~\ref{app:setup}. For each 1K steps, we estimate the NLL for validation. 

\subsection{Results}
Experimental results of Coupled-CVAE are shown in the main text.

\end{document}